\documentclass[letterpaper, 10 pt, conference]{ieeeconf}
\IEEEoverridecommandlockouts
\let\labelindent\relax
\usepackage{enumitem}
\usepackage{balance}
\usepackage[font={small}]{caption}
\usepackage{subcaption}
\usepackage{array}
\usepackage{textcomp}
\usepackage{mathtools, nccmath}
\usepackage{graphicx}
\usepackage{amsfonts}
\usepackage{amsmath}
\usepackage{amssymb}
\usepackage{algorithm}
\usepackage{algorithmic}
\usepackage{hyperref}
\usepackage{tikz}
\usepackage{arydshln}
\usepackage{multirow}
\usepackage{bm}
\usepackage{epstopdf}
\usepackage{cite}
\usepackage{siunitx}
\usepackage{multirow}
\usepackage{lipsum}

\usepackage{amsthm}

\newtheorem{theorem}{Theorem}

\newtheorem{remark}{Remark}
\theoremstyle{definition}
\newtheorem{definition}{Definition}
\newtheorem{problem}{Problem}
\newtheorem{assumption}{Assumption}

\title{\LARGE \bf
Safety-Critical Planning and Control for Dynamic Obstacle Avoidance
 Using Control Barrier Functions}

\author{Shuo Liu$^{*1}$, Yihui Mao$^{*1}$ and Calin A. Belta$^{2}$
\thanks{$^*$ Authors contributed equally.}
\thanks{This work was supported in part by the NSF under grant IIS-2024606 at Boston University.}
\thanks{$^{1}$S. Liu and Y. Mao are with the Department of Mechanical Engineering, Boston
University, Brookline, MA, USA. 
        {\tt\small \{liushuo, maoyihui\}@bu.edu}}%
\thanks{$^{2}$C. Belta is with the Department of Electrical and Computer Engineering and with the Department of Computer Science, University of Maryland, College Park, MD, USA 
        {\tt\small cbelta@umd.edu}}%
}

\begin{document} 
\maketitle
\begin{abstract}
Dynamic obstacle avoidance is a challenging topic for optimal control and optimization-based trajectory planning problems. Many existing works use Control Barrier Functions (CBFs). CBFs are typically formulated based on the distance to obstacles, or integrated with path planning algorithms as a safety enhancement tool. However, these approaches usually require knowledge of the obstacle boundary equations or have very slow computational efficiency. In this paper, we propose a framework based on model predictive control (MPC) with discrete-time high-order CBFs (DHOCBFs) to generate a collision-free trajectory. The DHOCBFs are first obtained from convex polytopes generated through grid mapping, without the need to know the boundary equations of obstacles. Additionally, a path planning algorithm is incorporated into this framework to ensure the global optimality of the generated trajectory. We demonstrate through numerical examples that our framework allows a unicycle robot to safely and efficiently navigate tight, dynamically changing environments with both convex and nonconvex obstacles. By comparing our method to established CBF-based benchmarks, we demonstrate superior computing efficiency, length optimality, and feasibility in trajectory generation and obstacle avoidance.
\end{abstract}

\section{Introduction}
\label{sec:Introduction}

The central problem in path planning is to determine a safe and efficient route for a robot to move from a start to a target, while avoiding obstacles \cite{gasparetto2015path}.
Sampling-based algorithms, such as Rapidly-exploring Random Trees (RRT) \cite{lavalle1998rapidly}, RRT Star (RRT*) \cite{karaman2011sampling} and Probabilistic Roadmaps (PRM) \cite{kavraki1996probabilistic}, explore feasible spaces randomly in high-dimensional environments, but their probabilistic completeness does not guarantee a fast solution. In contrast, grid-based search algorithms provide resolution completeness, ensuring rapid discovery of the shortest path between nodes. Algorithms like A* \cite{hart1968formal} combine actual and heuristic distances to find paths with the lowest estimated cost, guaranteeing optimality when heuristics are admissible. Jump Point Search (JPS) \cite{harabor2011online}, an optimized version of A*, accelerates searches in uniform grids by eliminating redundant nodes with jump points, effectively reducing computational cost and improving performance in large-scale spaces \cite{liu2017planning}.

Path planning algorithms often do not account for real-time changes, uncertainties, or dynamic obstacles in safety-critical environments, such as city highways with fast-moving vehicles and congested traffic. Control Barrier Functions (CBFs) are widely used to ensure safety in nonlinear systems by dynamically adjusting control inputs to maintain safe boundaries in real time \cite{ames2016control}. This approach breaks the timeline into brief intervals, solving a series of Quadratic Programs (QPs) at real-time speeds, effectively enforcing safety-critical constraints in applications like autonomous vehicles and bipedal robots \cite{ames2014control, hsu2015control}. 

Recently, CBFs have been applied to safety-critical path planners. In \cite{huang2023obstacle}, CBFs were combined with Control Lyapunov Functions (CLFs) in QPs for real-time navigation and dynamic collision avoidance. Other works integrated CBFs into the RRT or RRT* frameworks, enhancing obstacle avoidance and path optimality \cite{yang2019sampling, manjunath2021safe, ahmad2022adaptive}. Combining RRT*, discrete-time CBF, and Model Predictive Control (MPC) further improved safety by incorporating future state data \cite{jian2023dynamic}. To enhance computational efficiency and handle high relative degrees concerning system dynamics, discrete-time high-order CBFs (DHOCBFs) were proposed within an iterative MPC framework \cite{liu2023iterative}. However, these methods mainly focused on circular and elliptical obstacles, which often do not represent real-world scenarios. For complex shapes, polynomial CBFs using logistic regression were proposed in \cite{peng2023safe}, and linearized DHOCBFs predicted by a deep neural network were proposed in \cite{liu2024auxiliary}. However, these approaches are not suitable for dynamic obstacles.


We propose a novel framework for iterative MPC with DHOCBFs that detects complex-shaped obstacles using grid mapping. Unlike conventional approaches, our method leverages DHOCBFs extracted from convex polytopes in grid mapping, eliminating the need for precise boundary equations of obstacles. This enhances safety enforcement capabilities in densely clustered environments. Our method includes a novel optimal control framework specifically designed for dynamic obstacles, represented by occupied grids evolving over discrete time intervals. By incorporating a rapid path planning algorithm into the MPC, we generate optimal trajectories that adapt swiftly to changes in the environment. The use of DHOCBFs and system dynamics as linear constraints reformulates the MPC into a convex optimization problem, solved iteratively. Our approach ensures high-speed computation even in complex environments, demonstrating the efficiency of the framework.

We validate the effectiveness of our framework through extensive numerical examples. We show that it enables a unicycle robot to safely navigate through a densely clustered and dynamically changing map, effectively handling both convex and nonconvex obstacles. Our method not only provides faster control response times but also achieves higher feasibility rates and improved trajectory optimality compared to existing methods. This performance is particularly advantageous in complex scenarios where maintaining both rapid response and computational feasibility is crucial.

\section{Preliminaries}
\label{sec:Preliminaries}
In this section, we introduce some definitions and results
on Occupancy Grid Mapping (OGM) and DHOCBF.
\subsection{Occupancy Grid Mapping (OGM)}
\label{subsec:ogm}
Occupancy Grid Mapping (OGM), first proposed in \cite{moravec1985high}, is a method used in robotics and autonomous systems to represent the environment as a grid of cells. In this paper, there are three types of cells: fully occupied cells, partially occupied cells, and free cells.
\begin{definition}[Free Cell]
\label{def:free cell}
A free cell is defined as a grid cell that is not occupied by any obstacles.
\end{definition}
\begin{definition}[Occupied Cell]
\label{def:occupied cell}
An occupied cell is a grid cell that is fully or partially occupied by obstacles. If all adjacent cells are occupied, it is a fully occupied cell; if at least one adjacent cell is free, it is a partially occupied cell.
\end{definition}
Assume we set up a 2-D Cartesian coordinate system with the origin at 
$O$ to represent the environment. Each grid cell $n_{ij}$
is indexed by two integers, $i$ and $j$, representing the row and column indices, respectively. The cells are square-shaped with a constant side length of $d$, where the indices increase as positive integers along the positive axis and decrease as negative integers along the negative axis. The coordinates of each cell's geometric center, $(x_{ij},y_{ij})$, are defined by:
\begin{equation}
\label{eq:coordinate of cell}
\begin{split}
x_{ij}=id-sgn(i)\frac{d}{2}, y_{ij}=jd-sgn(j)\frac{d}{2},
\end{split}
\end{equation}
where $sgn(\cdot)$ extracts the sign of a real number. The resolution of obstacles depicted by grid cells can be improved by reducing the cell length $d.$

\subsection{Discrete-Time High-Order Control Barrier Function (DHOCBF)}
\label{subsec:dhocbf}

In this section, we consider a discrete-time control system in the form  
\begin{equation}
\label{eq:discrete-dynamics}
\mathbf{x}_{t+1}=f(\mathbf{x}_{t},\mathbf{u}_{t}),
\end{equation}
where $\mathbf{x}_{t} \in \mathcal X \subset \mathbb{R}^{n}$ represents its state at discrete time $t\in\mathbb{N}, \mathbf{u}_{t}\in \mathcal U \subset \mathbb{R}^{q}$ is the control input, and $f:\mathbb{R}^{n}\rightarrow\mathbb{R}^{q}$ is a locally Lipschitz function capturing the dynamics of the system.
We interpret safety as forward invariance of a set $\mathcal C$. A system is considered {\em safe} if, once it starts within set $\mathcal C$, it remains in $\mathcal C$ for all future times. We consider the set  $\mathcal C$ as the super-level set of a function $h: \mathbb{R}^{n}\to \mathbb{R}$:

\begin{equation}
\label{eq:safe-set}
\mathcal{C} \coloneqq \{\mathbf{x} \in \mathbb{R}^{n}: h(\mathbf{x} )\geq0 \}.
\end{equation}

\begin{definition}[Relative degree~\cite{sun2003initial}]
\label{def:relative-degree}
The output  $y_{t}=h(\mathbf{x}_{t})$ of system \eqref{eq:discrete-dynamics} is said to have relative degree $m$ with respect to dynamics (\ref{eq:discrete-dynamics}) if $\forall t\in\mathbb{N},$
\begin{equation}
\begin{split}
y_{t+i}=h(\bar{f}_{i-1}(f(\mathbf{x}_{t},\mathbf{u}_{t}))), \ i \in \{1,2,\dots,m\},\\
\text{s.t.} \ \frac{\partial y_{t+m}}{\partial \mathbf{u}_{t}}\ne \textbf{0}_{q}, \frac{\partial y_{t+i}}{\partial \mathbf{u}_{t}}=\textbf{0}_{q},  \ i \in \{1,2,\dots,m-1\},
\end{split}
\end{equation}
\end{definition}
where $\textbf{0}_{q}$ is the zero vector of dimension $q$. Informally, the relative degree $m$ is the number of steps (delay) in the output $y_{t}$ in order for the control input $\mathbf{u}_{t}$ to appear. In the above definition, we use $\bar{f}(\mathbf{x})$ to denote the uncontrolled state dynamics $f(\mathbf{x}, 0)$. The subscript $i$ of function $\bar{f}(\cdot)$ denotes the $i$-times recursive compositions of $\bar{f}(\cdot)$, i.e.,  $\bar{f}_{i}(\mathbf{x})=\underset{i\text{-times}~~~~~~~~~~~~~}{\underbrace{\bar{f}(\bar{f}(\dots,\bar{f}}(\bar{f}_{0}(\mathbf{x}))))}$ with $\bar{f}_{0}(\mathbf{x})=\mathbf{x}$.

We assume that $h(\mathbf{x})$ has
relative degree $m$ with respect to system (\ref{eq:discrete-dynamics}) based on Def. \ref{def:relative-degree}.
Starting with $\psi_{0}(\mathbf{x}_{t})\coloneqq h(\mathbf{x}_{t})$, we define a sequence of discrete-time functions $\psi_{i}:  \mathbb{R}^{n}\to\mathbb{R}$, $i=1,\dots,m$ as:
\begin{equation}
\label{eq:high-order-discrete-CBFs}
\psi_{i}(\mathbf{x}_{t})\coloneqq \bigtriangleup \psi_{i-1}(\mathbf{x}_{t},\mathbf{u}_{t})+\alpha_{i}(\psi_{i-1}(\mathbf{x}_{t})), 
\end{equation}
where $\bigtriangleup \psi_{i-1}(\mathbf{x}_{t}, \mathbf{u}_{t})\coloneqq \psi_{i-1}(\mathbf{x}_{t+1})-\psi_{i-1}(\mathbf{x}_{t})$, and $\alpha_{i}(\cdot)$ denotes the $i^{th}$ class $\kappa$ function which satisfies $\alpha_{i}(\psi_{i-1}(\mathbf{x}_{t}))\le \psi_{i-1}(\mathbf{x}_{t})$ for $i=1,\ldots, m$.
A sequence of sets $\mathcal {C}_{i}$ is defined based on \eqref{eq:high-order-discrete-CBFs} as
\begin{equation}
\label{eq:high-order-safety-sets}
\mathcal {C}_{i}\coloneqq \{\mathbf{x}\in \mathbb{R}^{n}:\psi_{i}(\mathbf{x})\ge 0\}, \ i =\{0,\ldots,m-1\}.
\end{equation}

\begin{definition}[DHOCBF \cite{xiong2022discrete}]
\label{def:high-order-discrete-CBFs}
Let $\psi_{i}(\mathbf{x}), \ i\in \{1,\dots,m\}$ be defined by \eqref{eq:high-order-discrete-CBFs} and $\mathcal {C}_{i},\ i\in \{0,\dots,m-1\}$ be defined by \eqref{eq:high-order-safety-sets}. A function $h:\mathbb{R}^{n}\to\mathbb{R}$ is a Discrete-Time High-Order Control Barrier Function (DHOCBF) with relative degree $m$ for system \eqref{eq:discrete-dynamics} if there exist $\psi_{m}(\mathbf{x})$ and $\mathcal {C}_{i}$ such that
\begin{equation}
\label{eq:highest-order-CBF}
\psi_{m}(\mathbf{x}_{t})\ge 0, \ \forall \mathbf{x}_{t}\in \mathcal{C}_{0}\cap \dots \cap \mathcal {C}_{m-1}, \forall t\in\mathbb{N}.
\end{equation}
\end{definition}

\begin{theorem}[Safety Guarantee \cite{xiong2022discrete}]
\label{thm:forward-invariance}
Given a DHOCBF $h(\mathbf{x})$  from Def. \ref{def:high-order-discrete-CBFs} with corresponding sets $\mathcal{C}_{0}, \dots,\mathcal {C}_{m-1}$ defined by \eqref{eq:high-order-safety-sets}, if $\mathbf{x}_{0} \in \mathcal {C}_{0}\cap \dots \cap \mathcal {C}_{m-1},$ then any Lipschitz controller $\mathbf{u}$ that satisfies the constraint in \eqref{eq:highest-order-CBF}, $\forall t\ge 0$ renders $\mathcal {C}_{0}\cap \dots \cap \mathcal {C}_{m-1}$ forward invariant for system \eqref{eq:discrete-dynamics}, $i.e., \mathbf{x}_{t} \in \mathcal {C}_{0}\cap \dots \cap \mathcal {C}_{m-1}, \forall t\ge 0.$
\end{theorem}

The function $\psi_{i}(\mathbf{x})$ in \eqref{eq:high-order-discrete-CBFs} is called a $i^{th}$ order Discrete-time Control Barrier Function (DCBF). Since satisfying the $i^{th}$ order DCBF constraint ($\psi_{i}(\mathbf{x}_{t})\ge0$) is a sufficient condition for rendering $\mathcal{C}_{0}\cap \dots \cap \mathcal{C}_{i-1}$ forward invariant for system \eqref{eq:discrete-dynamics} as shown above, it is not necessary to formulate DCBF constraints up to $m^{th}$ order as \eqref{eq:highest-order-CBF} if the control input $\mathbf{u}_{t}$ could be involved in some optimal control problem. In other words, the highest order for DCBF could be $m_{\text{cbf}}$ with $m_{\text{cbf}}\le m$ (see \cite{liu2023iterative}). In this paper, one simple method to define a $i^{th}$ order DCBF $\psi_{i}(\mathbf{x})$ in \eqref{eq:high-order-discrete-CBFs} is
\begin{equation}
\label{eq:simple-high-order-discrete-CBFs}
\psi_{i}(\mathbf{x}_{t})\coloneqq \bigtriangleup \psi_{i-1}(\mathbf{x}_{t},\mathbf{u}_{t})+\gamma_{i}\psi_{i-1}(\mathbf{x}_{t}),
\end{equation}
where $0<\gamma_{i}\le 1, i\in \{1,\dots,m_{\text{cbf}}\}$.
The expression in~\eqref{eq:simple-high-order-discrete-CBFs} follows the format of the first order DCBF proposed in \cite{agrawal2017discrete} and could be used to define a DHOCBF with arbitrary relative degree. 

In obstacle avoidance, $h$ is often defined by the distance to an obstacle's boundary, with states that keep the DHOCBF non-negative considered safe. However, complex obstacle shapes make their boundaries difficult to represent explicitly. In \cite{peng2023safe}, polynomial CBFs were derived from a grid map using logistic regression for static obstacles, but for dynamic obstacles, this approach would require continuous updates for polynomial CBFs, making it computationally intensive. The authors of \cite{liu2017planning} proposed using a safe flight corridor to obtain predefined convex overlapping polyhedra as a safe region from a grid map, without requiring the obstacle boundary equation. However, this method assumes static obstacles and lacks real-time adaptation to dynamic changes. In this paper, we demonstrate how to use polytopes to quickly partition the safe region among moving obstacles in grid maps and convert their edge equations into DHOCBFs to meet safety requirements.

\section{Problem Formulation and Approach}
\label{sec:Problem Formulation and Approach}


\begin{problem}
\label{prob:Path-prob}

Given an initial state $\mathbf{x}^{i}\in\mathbb{R}^{n}$, a final state $\mathbf{x}^{e}\in\mathbb{R}^{n}$, and a set of unsafe regions $\mathbb{O}_{i}, i\in \mathbb{I},\mathbb{O}_{i}\subset \mathbb{R}^{n}$, find a control strategy for system \eqref{eq:discrete-dynamics} that produces a trajectory starting at $\mathbf{x}^{i}$, ending at $\mathbf{x}^{e}$, avoiding the unsafe regions, and minimizing the energy of the controller. 
\end{problem}
It is important to note that, in the above problem, the unsafe regions may overlap, move, change shape, appear, or disappear at any time. If \eqref{eq:discrete-dynamics} represents the dynamics of a robot, the unsafe regions can be seen as configuration (C) - space obstacles, i.e., images of obstacles from workspace to configuration space. 

\begin{figure*}
\vspace{3mm}
    \centering
    \includegraphics[scale=0.33]{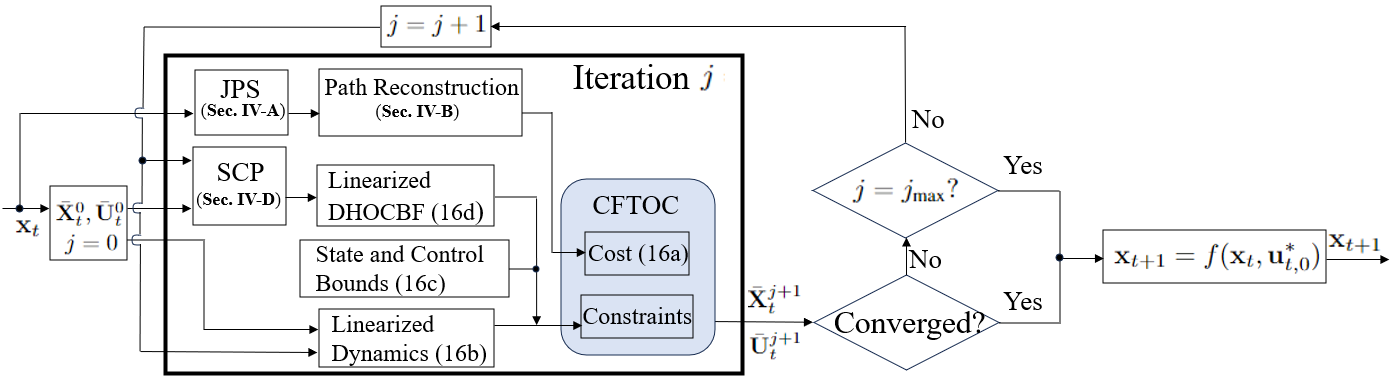}
    \caption{Schematic of the iterative process of solving the convex MPC at time $t$.}
    \label{fig:iteration-module}
\end{figure*}

\textbf{Approach:} Our proposed approach, illustrated in Fig. \ref{fig:iteration-module}, combines iterative MPC-DHOCBF (iMPC-DHOCBF \cite{liu2023iterative}) with JPS and Safe Convex Polytopes (SCP) tailored for dynamic grid mapping. In this work, obstacles are detected using OGM, as discussed in Sec. \ref{subsec:ogm}. Since the obstacles are moving, we capture them with grids at each discrete time step $t\in \mathbb{N}$, where $t$ corresponds to the discretization time of the system \eqref{eq:discrete-dynamics}. The set of partially occupied cells, $\mathbb{G}_{t}\subset \mathbb{R}^{n}$, represents obstacle boundaries.

At $t=0$, starting from the initial state $\mathbf{x}^{i}$ (with $\mathbf{x}_{0}=\mathbf{x}^{i},\mathbf{x}_{e}=\mathbf{x}^{e}$), we identify the cells (geometry centers) closest to the position components of $\mathbf{x}_{0}$ and $\mathbf{x}_{e}$, denoted as $n_{0}$ and $n_{e}$. JPS is then used to find a safe path $\pi_{0} = \left \langle n_{0}, \ldots, n_{e} \right \rangle$, and a path reconstruction approach converts the cells along the path into reference states $\mathbf{X}_{r,0}=\left \langle \mathbf{x}_{r,0}, \ldots, \mathbf{x}_{r,e} \right \rangle,$ serving as waypoints for the system to follow.

However, since the safe path found by JPS is based solely on the current time $t$, it may intersect with dynamic obstacles at future times $t+i,1 \le i\le N$. To ensure safety throughout the path-following phase, we utilize DHOCBFs. SCPs are used to obtain safe regions geometrically by arranging the partially occupied cells $G_{0}$ based on their geometric distance from $n_{0}$, connecting each to $n_{0}$ with line segments, and drawing perpendicular lines through them until a closed polytope is formed (see Sec. \ref{subsec:iMPCSCP}). The edges of this polytope are used to derive linear DHOCBFs, eliminating the need for explicit obstacle boundary equations.

The above process is repeated iteratively for all future time steps 
$t\le T$, under the framework of iterative MPC. At each time step, the cost function over a receding horizon $N\le T$ is minimized:
 \begin{small}
\begin{equation}
\label{eq:cost-function-1}
\begin{split}
J(\mathbf{u}_{t,k})=\sum_{k=0}^{N-1}\mathbf{u}_{t,k}^{T}\mathbf{u}_{t,k}+\sum_{k=0}^{N}(\mathbf{x}_{t,k}-\mathbf{x}_{r,k})^{T}(\mathbf{x}_{t,k}-\mathbf{x}_{r,k}),
\end{split}
\end{equation}
 \end{small}
where $\mathbf{x}_{t, k}$, $\mathbf{u}_{t, k}$ are the state and input predictions of system \eqref{eq:discrete-dynamics} at time $t+k$ made at the current time $t$, with $\mathbf{x}_{t, 0}=\mathbf{x}_{t}, \mathbf{u}_{t, 0}=\mathbf{u}_{t}.$ Linear DHOCBFs are obtained at each time step, serving as safety constraints to ensure obstacle avoidance. The reference states $\mathbf{X}_{r,t}$ are used in \eqref{eq:cost-function-1} to guide the system from $\mathbf{x}_{t}$ to $\mathbf{x}_{e}$, minimizing the energy cost \eqref{eq:cost-function-1}.

\section{Iterative MPC with DHOCBF in Dynamic Grid Mapping}
\label{sec:Methodology}

We use the optimal values of states $\mathbf{X}_{t-1}^{*}$ and inputs  $\mathbf{U}_{t-1}^{*}$ obtained at time $t-1$ as the nominal states  and inputs for iteration 0 at the current time step $t$, denoted as $\mathbf{X}_{t-1}^{*}=\bar{\mathbf{X}}_{t}^{0}, \mathbf{U}_{t-1}^{*}=\bar{\mathbf{U}}_{t}^{0}$ (if $t=0$, the initial guess for the control input could be $\bar{\mathbf{U}}_{0}^{0}=0,$ based on which we get the initial nominal state $\bar{\mathbf{X}}_{0}^{0}$).
In each iteration $j$, we solve a convex finite-time optimal control (CFTOC) problem with linearized dynamics and linear DHOCBFs to get optimal values of states and inputs $\mathbf{X}_{t}^{*, j}=[\mathbf{x}_{t, 0}^{*, j},\dots, \mathbf{x}_{t, N}^{*, j}], \mathbf{U}_{t}^{*, j}=[\mathbf{u}_{t,0}^{*, j},\dots, \mathbf{u}_{t,N-1}^{*, j}],$ then update the state and input vectors for next iteration, i.e., $\bar{\mathbf{X}}_{t}^{j+1} = \mathbf{X}_{t}^{*, j}, \bar{\mathbf{U}}_{t}^{j+1} = \mathbf{U}_{t}^{*, j}.$ The iteration for the current time step is completed if the convergence error is within a user-defined normalized convergence criterion or if the iteration count reaches the maximum allowable number $j_{\text{max}}$. The optimized states $\mathbf{X}_{t}^{*}$ and inputs $\mathbf{U}_{t}^{*}$ are then passed to the CFTOC formulation for the next time step. Notice that the open-loop trajectory with updated states $\bar{\mathbf{X}}_{t}^{j} = [\bar{\mathbf{x}}_{t, 0}^{j},\dots, \bar{\mathbf{x}}_{t, N-1}^{j}]$ and inputs $\bar{\mathbf{U}}_{t}^{j} = [\bar{\mathbf{u}}_{t, 0}^{j},\dots, \bar{\mathbf{u}}_{t, N-1}^{j}]$ is passed between iterations, which allows iterative linearization for system dynamics locally. At each time step, the state of the output closed-loop trajectory is updated by $\mathbf{x}_{t+1} = f(\mathbf{x}_{t}, \mathbf{u}_{t,0}^{*})$. 

\subsection{Dynamic Path Planning-JPS}
\label{subsec:DPP}
As the system moves, the locations of dynamic obstacles change as well, highlighting the necessity of path planning at each time step (dynamic path planning). In Fig. \ref{fig:iteration-module}, the system state $\mathbf{x}_{t}$ is updated by solving the CFTOC at time step $t-1$. The current and final position components of $\mathbf{x}_{t}$ and $\mathbf{x}_{e}$ are $(x_{t},y_{t})$ and $ (x_{e},y_{e})$. Based on the OGM discussed in Sec. \ref{subsec:ogm}, we search for free cells whose geometric center coordinates are closest to $(x_{t},y_{t})$ and $(x_{e},y_{e})$, denoted as $n_{t}$ and $n_{e},$ respectively. The optimal safe path is then determined by JPS as $\pi_{t}=\left \langle n_{t}, n_{t+1},\dots,n_{e} \right \rangle$ consisting of a sequence of free cells coordinates from $n_{t}$ to $n_{e}.$ Following this path ensures safe and efficient navigation for the system.

\subsection{Path Reconstruction}
\label{subsec:RP}

To illustrate the necessity of path reconstruction, consider a simplified unicycle model in the form
\begin{equation}
\label{eq:simplified-unicycle-model}
\begin{bmatrix} x_{t+1}-x_{t} \\ y_{t+1}-y_{t} \\ \theta_{t+1}-\theta_{t} \\ v_{t+1}-v_{t} \end{bmatrix}=\begin{bmatrix} v_{t}\cos(\theta_{t}) \Delta t \\ v_{t}\sin(\theta_{t}) \Delta t \\ 0 \\0 \end{bmatrix}+\begin{bmatrix} 0  \\ 0 \\ \Delta t \\0  \end{bmatrix}
 u_{t},
\end{equation}
where $\mathbf{x}_{t}=[x_{t},y_{t},\theta_{t},v_{t}]^{T}$ captures the 2-D location and heading angle, $v_{t}$ denotes constant linear speed, and $\mathbf{u}_{t}$ represents angular velocity. $\Delta t$ denotes the discretized time interval. Note that the reference state in the cost \eqref{eq:cost-function-1} for this system is $\mathbf{x}_{r,t}=[x_{r,t},y_{r,t},\theta_{r,t},v_{r,t}]^{T}.$ However, when using the cells (waypoints) of the safe path $\pi_{t}$ in Sec. \ref{subsec:DPP} as the reference, each cell on the path is defined only by the coordinate $(x,y)$, meaning the desired heading angle and speed $\theta_{r},v_{r}$ are not specified. To address this problem, we make the following assumption: 

\begin{assumption}\label{asm:Feasibility Constraint}
Given a discrete-time control system \eqref{eq:discrete-dynamics}, where the state $\mathbf{x}_{t}$ includes the position components at each time step $t, \forall t\in\mathbb{N}$, specifically 
$(x_{t},y_{t})$ for 2-D or $(x_{t},y_{t},z_{t})$ for 3-D, the remaining components of the state can always be determined based on the position components.
\end{assumption}

Revisiting system \eqref{eq:simplified-unicycle-model} as a special case for Assumption 1, the heading angle and linear speed at $t$ are given by:
\begin{equation}
\label{eq:heading angle1}
\theta_{t}=\arctan(\frac{y_{t+1}-y_{t}}{x_{t+1}-x_{t}}), v_{t}=\frac{x_{t+1}-x_{t}}{\cos(\theta_{t}) \Delta t}.
\end{equation}
If the waypoints starting from $t$ are found by JPS, i.e., the path $\pi_{t}=\left \langle n_{t}, n_{t+1},\dots,n_{e} \right \rangle$ is found, we use the coordinates of each waypoint $n_{t+k}$ as the reference location $(x_{r,t+k},y_{r,t+k})$ for the specific time step $t+k$, where $0\le k\le N$. Then, based on Assumption 1, the reference state $\mathbf{x}_{r,t+k}$ can be determined for this time step. The reference states $\mathbf{X}_{r,t}=\left \langle \mathbf{x}_{r,t}, \mathbf{x}_{r,t+1},\dots,\mathbf{x}_{r,t+N} \right \rangle$ are used in the cost \eqref{eq:cost-function-1} to minimize the difference between the system's location and the reference location over a given horizon. This process is referred to as path reconstruction in this paper.
\begin{remark}
\label{rem:waypoints length} 
The distance between two adjacent waypoints on the path $\pi_{t}$ may be too large to follow, e.g., the reference velocity $v_{t}$ defined by $\frac{x_{t+1}-x_{t}}{\cos(\theta_{t}) \Delta t}$ in \eqref{eq:heading angle1} may be unattainable for system \eqref{eq:simplified-unicycle-model} because of the input constraints. To address this, additional waypoints can be inserted along the line segment between adjacent waypoints to reduce their distance. If the waypoints on the reference path $\pi_{t}$ are very close to the obstacle, closely following its boundary, the risk of collision increases. To mitigate this, we can use the distance map planner from \cite{liu2017planning}, which utilizes an artificial potential field to keep the waypoints at a safe distance from the obstacle. This method balances the safety and length optimality of the generated reference path and is applied in Sec. \ref{sec:Case Study and Simulations}.
\end{remark}

\subsection{Linearization of Dynamics}
\label{subsec:linearization-dynamics}
At iteration $j$, an improved vector $\mathbf{u}_{t,k}^{j}$ is achieved by linearizing the system around $\bar{\mathbf{x}}_{t,k}^{j}, \bar{\mathbf{u}}_{t,k}^{j}$:
\begin{equation}
\begin{split}
\label{eq:linearized-dynamics}
\mathbf{x}_{t,k+1}^{j}{-}\bar{\mathbf{x}}_{t,k+1}^{j}{=}A^{j}(\mathbf{x}_{t,k}^{j}{-}\bar{\mathbf{x}}_{t,k}^{j})+B^{j}(\mathbf{u}_{t,k}^{j}{-}\bar{\mathbf{u}}_{t,k}^{j}),
\end{split}
\end{equation}
where $0 \leq j < j_{\text{max}}$; $k$ and $j$ represent open-loop time step and iteration indices, respectively. We also have
\begin{equation}
A^{j}=D_{\mathbf{x}}f(\bar{\mathbf{x}}_{t,k}^{j}, \bar{\mathbf{u}}_{t,k}^{j}), \ B^{j}=D_{\mathbf{u}}f(\bar{\mathbf{x}}_{t,k}^{j}, \bar{\mathbf{u}}_{t,k}^{j}),
\end{equation}
where $D_{\mathbf{x}}$ and $D_{\mathbf{u}}$ denote the Jacobian of the system dynamics \eqref{eq:discrete-dynamics} with respect to the state $\mathbf{x}$ and the input $\mathbf{u}$, respectively.
This approach allows to linearize the system at $(\bar{\mathbf{x}}_{t,k}^{j}, \bar{\mathbf{u}}_{t,k}^{j})$ locally between iterations.
The convex system dynamics constraints are given in \eqref{eq:linearized-dynamics},  with all nominal vectors 
$(\bar{\mathbf{x}}_{t,k}^{j}, \bar{\mathbf{u}}_{t,k}^{j})$ from the current iteration treated as constants, constructed from the previous iteration $j-1$.
 \subsection{DHOCBF from Safe Convex Polytope}
\label{subsec:iMPCSCP}
\begin{figure}[ht]
    \centering
    \includegraphics[scale=0.33]{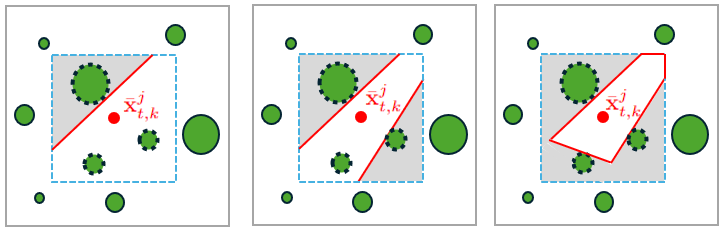}
    \caption{Schematic illustrating how to find the SCP (red). The obstacle detection range is denoted by a blue dashed square centered at $\bar{\mathbf{x}}_{t, k}^{j}$. The obstacles are depicted by green circles, with their boundaries captured by partially occupied cells, and the centers of the cells are marked by black points.}
    \label{fig:SCP}
\end{figure}

 The Safe Convex Polytope (SCP) outlines the safe region, with its boundaries (linear equations) used to formulate DHOCBFs. To construct the SCP, based on the grid maps $M_{t+k}, 0\le k\le N$, at iteration $j$, a square (depicted by the blue square with dashed lines in Fig. \ref{fig:SCP}) delineating the obstacle detection range is drawn, with its geometric center at the current location of the system ($\bar{\mathbf{x}}_{t, k}^{j}$). Obstacles within this square are detected by identifying partially occupied cells $\mathbb{G}_{t}\subset \mathbb{R}^{n}$. The cell from $\mathbb{G}_{t}$ closest to the system is identified, and a perpendicular line is drawn from this cell to the line connecting the cell and the system, forming the first edge of the SCP. This cell is then excluded from $\mathbb{G}_{t}$. The area opposite the system is marked as dangerous (marked gray in Fig. \ref{fig:SCP}) and also excluded. This process repeats for undetected areas within the square until the closed SCP boundary (a polytope) is established. This boundary consists of $n_{t,k}^{j}$ line segments (edges) and each line can be expressed by a linear equation relevant to $\bar{\mathbf{x}}_{t, k}^{j}$ as $h_{l}(\mathbf{x}_{t, k}^{j},\bar{\mathbf{x}}_{t, k}^{j})$, where $l$ represents the index of the edge and $l\le n_{t,k}^{j}.$ This allows us to define a safe region (inner area of the polytope) expressed by ${C}_{t, k}^{j}\coloneqq \{\mathbf{x}_{t, k}^{j}\in \mathbb{R}^{n}:h_{\text{scp}}(\mathbf{x}_{t, k}^{j},\bar{\mathbf{x}}_{t, k}^{j}) \ge 0\}, h_{\text{scp}}(\mathbf{x}_{t, k}^{j},\bar{\mathbf{x}}_{t, k}^{j})\ge0 \Longleftrightarrow   h_{1}(\mathbf{x}_{t, k}^{j},\bar{\mathbf{x}}_{t, k}^{j})\ge0 \cap \dots \cap h_{n_{t,k}^{j}}(\mathbf{x}_{t, k}^{j},\bar{\mathbf{x}}_{t, k}^{j}) \ge 0 .$ The relative degree of $h_{l}(\mathbf{x}_{t, k}^{j},\bar{\mathbf{x}}_{t, k}^{j})$ with respect to system \eqref{eq:discrete-dynamics} is $m.$

In order to guarantee safety with forward invariance based on Thm. \ref{thm:forward-invariance}, we derive a sequence of DHOCBFs up to the order $m_{\text{cbf}}$:
\begin{small}
\begin{equation}
\label{eq:DCBFs}
\begin{split}
 \tilde{\psi}_{0}^{l}(\mathbf{x}_{t,k}^{j}) \coloneqq & h_{l}(\mathbf{x}_{t, k}^{j},\bar{\mathbf{x}}_{t, k}^{j}) \\
 \tilde{\psi}_{i}^{l}(\mathbf{x}_{t,k}^{j}) \coloneqq & \tilde{\psi}_{i-1}^{l}(\mathbf{x}_{t,k+1}^{j})-\tilde{\psi}_{i-1}^{l}(\mathbf{x}_{t,k}^{j}){+}\gamma_{i}^{l}\tilde{\psi}_{i-1}^{l}(\mathbf{x}_{t,k}^{j}),
 \end{split}
\end{equation}
\end{small}
where $0 <\gamma_{i}^{l} \le 1$, $i\in\{1,\dots,m_{\text{cbf}}\}, l\in \{1,\dots,n_{t,k}^{j}\},$ and $m_{\text{cbf}} \le m$ (as in \eqref{eq:simple-high-order-discrete-CBFs}).

 As the number of obstacles within the obstacle detection
range increases, $n_{t,k}^{j}$ may become very large. This leads to a high number of safety constraints, which may make solving the optimization problem infeasible. In order to handle this issue, we introduce a slack variable $\omega_{t,k,i}^{j,l}$ with a corresponding decay rate $(1-\gamma_{i}^{l}).$ Similar to \cite{liu2023iterative}, we replace $\tilde{\psi}_{i}^{l}(\mathbf{x}_{t,k}^{j})\ge 0$ in \eqref{eq:DCBFs} with
\begin{equation}
\label{eq:convex-hocbf-constraint}
\begin{split}
& \tilde{\psi}_{i-1}^{l}(\mathbf{x}_{t,k}^{j}) + \sum_{\nu =1}^{i}Z_{\nu,i}^{l}(1-\gamma_{i}^{l})^{k}\tilde{\psi}_{0}^{l}(\mathbf{x}_{t,\nu}^{j})\ge \\
& \quad \omega_{t,k,i}^{j,l} Z_{0,i}^{l}(1-\gamma_{i}^{l})^{k}\tilde{\psi}_{0}^{l}(\mathbf{x}_{t,0}^{j}), j \le j_{\text{max}} \in \mathbb{N}^{+},\\
&  l\in \{1,\dots,n_{t,k}^{j}\}, \ i\in \{1,\dots,m_{\text{cbf}}\},  \ \omega_{t,k,i}^{j,l}\in  \mathbb{R},
\end{split}
\end{equation}
where $Z_{\nu,i}^{l}$ is a constant that aims to make constraint \eqref{eq:convex-hocbf-constraint} linear in terms of decision variables $\mathbf{x}_{t,k}^{j}, \omega_{t,k,i}^{j,l},$ and can be obtained in [Eqn. (15), \cite{liu2023iterative}].


\subsection{CFTOC Problem}
\label{subsec:convex-mpc-dhocbf}
\begin{figure*}[ht]
\vspace{3mm}
    \centering
    \includegraphics[scale=0.36]{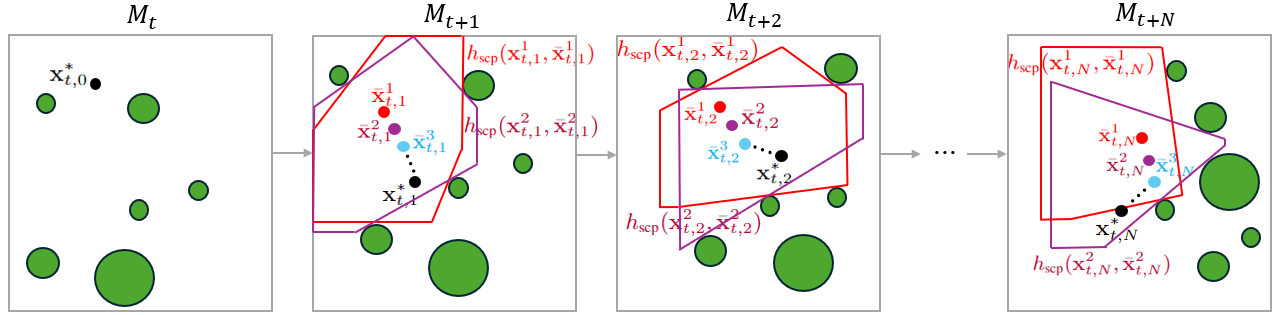}
    \caption{Finding the open-loop trajectory (black points) based on SCP in grid maps starting at $t$. Polytopes of each color are generated based on the points of the corresponding color, and points in the next iteration $(\bar{\mathbf{x}}_{t,k}^{j+1})$ are confined and generated within the polytopes $h_{\text{scp}}(\mathbf{x}_{t, k}^{j},\bar{\mathbf{x}}_{t, k}^{j})$ created at the current iteration. The point $(\mathbf{x}_{t,1}^{*})$ is selected as the starting point for solving the CFTOC starting at $t+1$.}
    \label{fig:rapid algorithm}
\end{figure*}
\noindent\rule{\columnwidth}{0.4pt}
  \textbf{CFTOC of iMPC-DHOCBF at iteration $j$:}  
\begin{subequations}
{\small
\label{eq:impc-dcbf}
\begin{align}
\label{eq:impc-dcbf-cost}
  & \min_{\mathbf{U}_{t}^{j},\Omega_{t,1}^{j,l},\dots, \Omega_{t,m_{\text{cbf}}}^{j,l}} p(\mathbf{x}_{t,N}^{j})+\sum_{k=0}^{N-1} q(\mathbf{x}_{t,k}^{j},\mathbf{u}_{t,k}^{j},\omega_{t,k,i}^{j,l}) \\
   \text{s.t.} \ & \mathbf{x}_{t,k+1}^{j}{-} \bar{\mathbf{x}}_{t,k+1}^{j}{=}A^{j}(\mathbf{x}_{t,k}^{j}-\bar{\mathbf{x}}_{t,k}^{j}){+}B^{j}(\mathbf{u}_{t,k}^{j}-\bar{\mathbf{u}}_{t,k}^{j}), \label{subeq:impc-dcbf-linearized-dynamics}\\
    & \mathbf{u}_{t,k}^{j} \in \mathcal U, \  \mathbf{x}_{t,k}^{j} \in \mathcal X, \ \omega_{t,k,i}^{j,l}\in \mathbb{R},\label{subeq:impc-dcbf-variables-bounds}\\
    & \tilde{\psi}_{i-1}^{l}(\mathbf{x}_{t,k}^{j})+ \sum_{\nu=1}^{i}Z_{\nu,i}^{l}  (1-\gamma_{i}^{l})^{k}\tilde{\psi}_{0}^{l}(\mathbf{x}_{t,\nu}^{j})  \ge \nonumber \\ 
    & \omega_{t,k,i}^{j,l}Z_{0,i}^{l}(1-\gamma_{i}^{l})^{k}\tilde{\psi}_{0}^{l}(\mathbf{x}_{t,0}^{j}) \label{subeq:impc-dcbf-linearized-hocbf},
\end{align}
}
\end{subequations}
\noindent\rule{\columnwidth}{0.4pt}

In Secs.~\ref{subsec:linearization-dynamics} and \ref{subsec:iMPCSCP}, we have illustrated the linearization of system dynamics as well as the process to get linear DHOCBFs. This approach allows us to incorporate these constraints within a convex MPC framework at each iteration, a method we refer to as convex finite-time constrained optimization control (CFTOC). This is solved at iteration $j$ with optimization variables $\mathbf{U}_{t}^{j} = [ \mathbf{u}_{t,0}^{j}, \dots, \mathbf{u}_{t,N-1}^{j}]$ and $\Omega_{t,i}^{j,l} = [\omega_{t,0,i}^{j,l},\dots, \omega_{t,N,i}^{j,l}]$, where $i\in \{1,\dots, m_{\text{cbf}}\}, l\in \{1,\dots, n_{t,k}^{j}\}.$ 

In the CFTOC, the linearized dynamics constraints in~\eqref{eq:linearized-dynamics} and the DHOCBF constraints in~\eqref{eq:convex-hocbf-constraint} are enforced using constraints~\eqref{subeq:impc-dcbf-linearized-dynamics} and ~\eqref{subeq:impc-dcbf-linearized-hocbf} at each open loop time step $k\in\{0,\dots, N-1\}$.
State and input constraints are considered in~\eqref{subeq:impc-dcbf-variables-bounds}. 
 Slack variables remain unconstrained as the optimization aims to minimize deviation from nominal DHOCBF constraints through the cost term $q(\cdot, \cdot, \omega_{t,k,i}^{j,l})$, ensuring feasibility, as discussed in~\cite{liu2023iterative}.
To maintain the safety guarantee from the DHOCBFs, constraints~\eqref{subeq:impc-dcbf-linearized-hocbf} are strictly enforced for $i\in\{0,\dots, m_{\text{cbf}}\}, l\in \{1,\dots, n_{t,k}^{j}\}.$ 
The optimal decision variables at iteration $j$ include the control input vectors $\mathbf{U}_{t}^{*,j}=[\mathbf{u}_{t,0}^{*,j},\dots,\mathbf{u}_{t,N-1}^{*,j}]$ and slack variable vectors $\mathbf{\Omega}_{t,i}^{*,j}=[\Omega_{t,i}^{*,j,1},\dots,\Omega_{t,i}^{*,j, n_{t,k}^{j}}]$ with $\Omega_{t, i}^{*,j,l} = [\omega_{t,0,i}^{*,j,l},\dots, \mathbf{\omega}_{t,N,i}^{*,j,l}].$ 
The CFTOC is solved iteratively in our proposed iMPC-DHOCBF with the grid maps over horizon $N$ as $[M_{t}, M_{t+1},\dots,M_{t+N}].$ In Fig.~\ref{fig:rapid algorithm}, red polytopes are generated based on red points $(\bar{\mathbf{x}}_{t,k}^{1})$, and purple points $(\bar{\mathbf{x}}_{t,k}^{2})$ are confined within the red polytopes in which they are generated. Again, blue points $(\bar{\mathbf{x}}_{t,k}^{3})$ are confined and genereted in pink polotopes. This process repeats until convergence criteria or the maximum iteration number $j_{\text{max}}$ is reached, allowing extraction of the safe open-loop trajectory (black points $[\mathbf{x}_{t,0}^{*},\mathbf{x}_{t,1}^{*},\dots, \mathbf{x}_{t,N}^{*}]$). 

\section{Case Study and Simulations}
\label{sec:Case Study and Simulations}
In this section, we present numerical results to demonstrate the effectiveness of our proposed method by applying it to a unicycle model, and by comparing it to established CBF-based benchmarks.
For iMPC-DHOCBF, we used OSQP~\cite{stellato2020osqp} to solve the convex optimizations at all iterations. The grid mapping simulator was built using RViz in the ROS Noetic. We used a Linux desktop with Intel Core i9-13900H running c++ for all computations. The animation video can be found at \url{https://youtu.be/iiihs_KPEZQ}.
\subsection{Numerical Setup}

\subsubsection{System Dynamics}
Consider a discrete-time unicycle model in the form
\begin{equation}
\label{eq:unicycle-model}
\begin{bmatrix} x_{t+1}{-}x_t \\ y_{t+1}{-}y_t \\ \theta_{t+1}{-}\theta_t \\ v_{t+1}{-}v_t \end{bmatrix}{=}\begin{bmatrix} v_{t} \cos(\theta_{t}) \Delta t \\ v_{t}\sin(\theta_{t}) \Delta t \\ 0 \\ 0 \end{bmatrix}{+}\begin{bmatrix} 0 & 0 \\ 0 & 0 \\ \Delta t & 0 \\ 0 & \Delta t \end{bmatrix}
\begin{bmatrix} u_{1,t} \\ u_{2,t} \end{bmatrix},
\end{equation}
where $\mathbf{x}_{t}=[x_{t},y_{t},\theta_{t},v_{t}]^{T}$ captures the 2-D location, heading angle, and linear speed; $\mathbf{u}_{t}=[u_{1,t},u_{2,t}]^{T}$ represents angular velocity ($u_{1}$) and linear acceleration ($u_{2}$), respectively.
The system is discretized with $\Delta t = 0.01$ and $T=3000$. System~\eqref{eq:unicycle-model} is subject to the following constraints:
\begin{equation}
\begin{split}
\label{eq:state-input-constraint}
\mathcal{X}=\{\mathbf{x}_{t}\in \mathbb{R}^{4}:& [-50, -50,-10,-30]^{T} \le \mathbf{x}_{t},\\
& \mathbf{x}_{t} \le [50, 50,10,30]^{T}\},\\
\mathcal{U}=\{\mathbf{u}_{t}\in \mathbb{R}^{2}:& [-15, -5]^{T} \le \mathbf{u}_{t}\le [15, 5]^{T}\}.
\end{split}
\end{equation}

\subsubsection{System Configuration}
The initial state is $[-45,-45,\frac{\pi}{4},25]^{T},$ the goal location is $[45, 45]^{T},$ and the reference states at time step $t$ up to horizon $N$ are $\mathbf{x}_{r,t+k}=[x_{r,t+k},y_{r,t+k},\theta_{r,t+k},v_{r,t+k}]^{T}, k\in\left \{0,\dots,N \right \},$ determined by the path reconstruction in Sec. \ref{subsec:RP}, with $v_{r,t+k}=25.$ The other reference vectors are $\mathbf{u}_{r}=[0,0]^{T}$ and $\omega_{r}=[1,1]^{T}$. The robot will stop once it reaches within a 0.1 radius of the goal location.

\subsubsection{DHOCBF}
We get each candidate DHOCBF $h_{l}(\mathbf{x}_{t, k}^{j},\bar{\mathbf{x}}_{t, k}^{j})$ from SCP introduced in Sec. \ref{subsec:iMPCSCP} and set $m_{\text{cbf}}=1.$ In \eqref{eq:convex-hocbf-constraint}, we have $Z_{0,1}^{l}=1$ and $\gamma_{1}^{l}=0.1.$ 
The obstacle detection range, represented by the side length of the blue dashed square in Fig. \ref{fig:SCP}, is 40.

\subsubsection{MPC Design}
The cost function of the MPC problem consists of stage cost
$q(\mathbf{x}_{t,k}^j,\mathbf{u}_{t,k}^{j},\omega_{t,k,i}^{j,l})= \sum_{k=0}^{N-1} (||\mathbf{x}_{t,k}^{j}-\mathbf{x}_{r,t+k}||_Q^2 + ||\mathbf{u}_{t,k}^{j}-\mathbf{u}_{r}||_R^2 +||\omega_{t,k,i}^{j,l}-\omega_{r}||_S^2)$
and terminal cost $p(\mathbf{x}_{t,N}^{j})=||\mathbf{x}_{t,N}^{j}-\mathbf{x}_{r,t+N}||_P^2$, where $Q=P=[10000, 10000,100,10]^{T}, R= \mathcal{I}_{2}$ and $S=100000\cdot \mathcal{I}_{2}$.

\subsubsection{Convergence Criteria}
We use the following absolute and relative convergence functions as convergence criteria mentioned in Fig. \ref{fig:iteration-module}:
\begin{equation}
\label{eq:convergence-criteria}
    \begin{split}
    e_{\text{abs}}(\mathbf{X}_{t}^{*, j}, \mathbf{U}_{t}^{*, j}) &= ||\mathbf{X}^{*, j} - \bar{\mathbf{X}}^{*, j}|| \\
    e_{\text{rel}}(\mathbf{X}_{t}^{*, j}, \mathbf{U}_{t}^{*, j}, \bar{\mathbf{X}}_{t}^{j}, \bar{\mathbf{U}}_{t}^{j}) &= ||\mathbf{X}^{*, j} - \bar{\mathbf{X}}^{*, j}||/||\bar{\mathbf{X}}^{*, j}||.
\end{split}
\end{equation}
The iterative optimization stops when $e_{\text{abs}} < \varepsilon_{\text{abs}}$ or $e_{\text{rel}} < \varepsilon_{\text{rel}}$, where $\varepsilon_{\text{abs}} = 0.05$, $\varepsilon_{\text{rel}} = 10^{-2}$ and the maximum iteration number is set as $j_{\text{max}} = 10$.

\subsection{Performance}

\subsubsection{Avoiding Convex Dynamic Obstacles}
\begin{figure*}[t]
\vspace{3mm}
\centering
\begin{subfigure}{.160\textwidth}
  \centering
  \includegraphics[width=\linewidth]{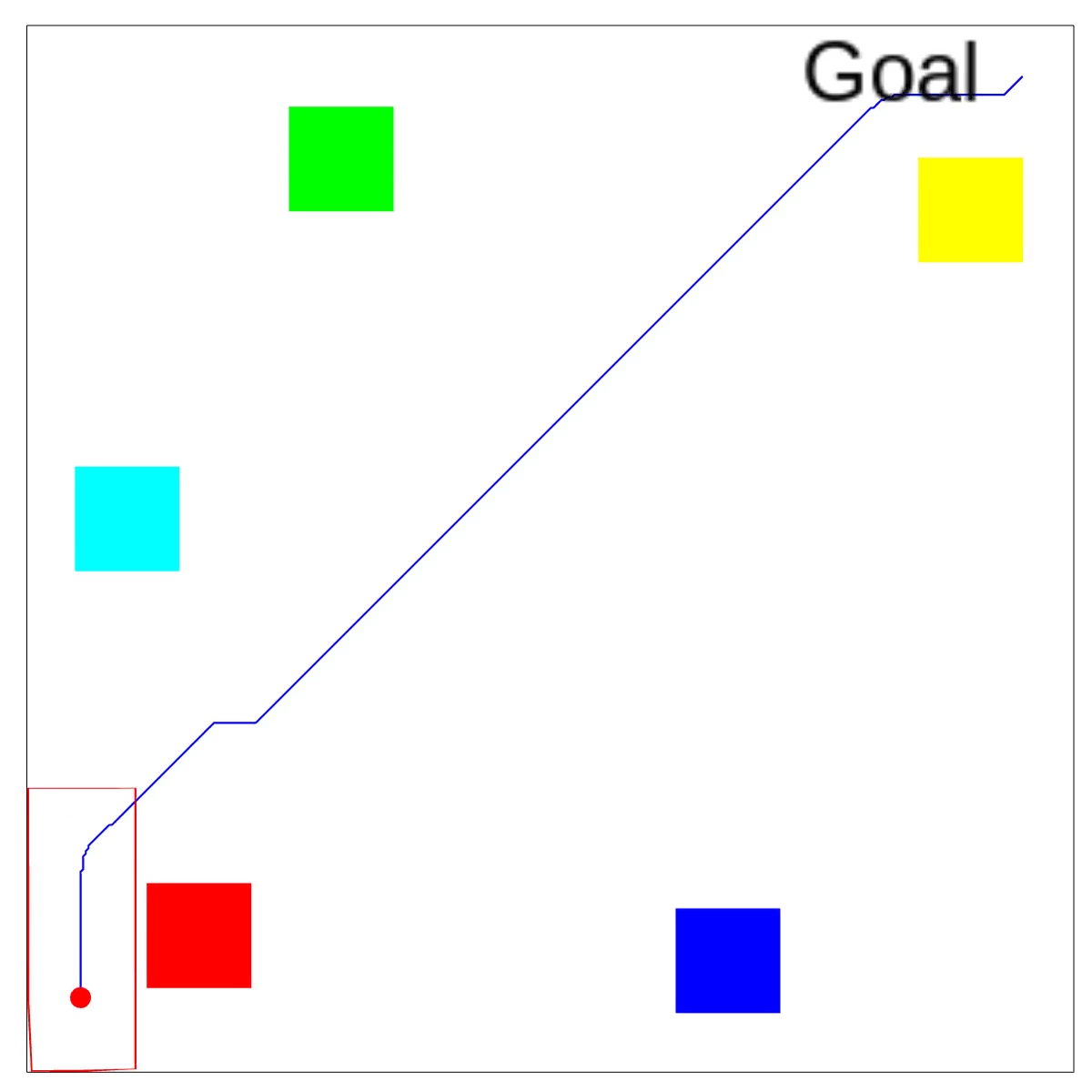}
\end{subfigure}
\begin{subfigure}{.160\textwidth}
  \centering
  \includegraphics[width=\linewidth]{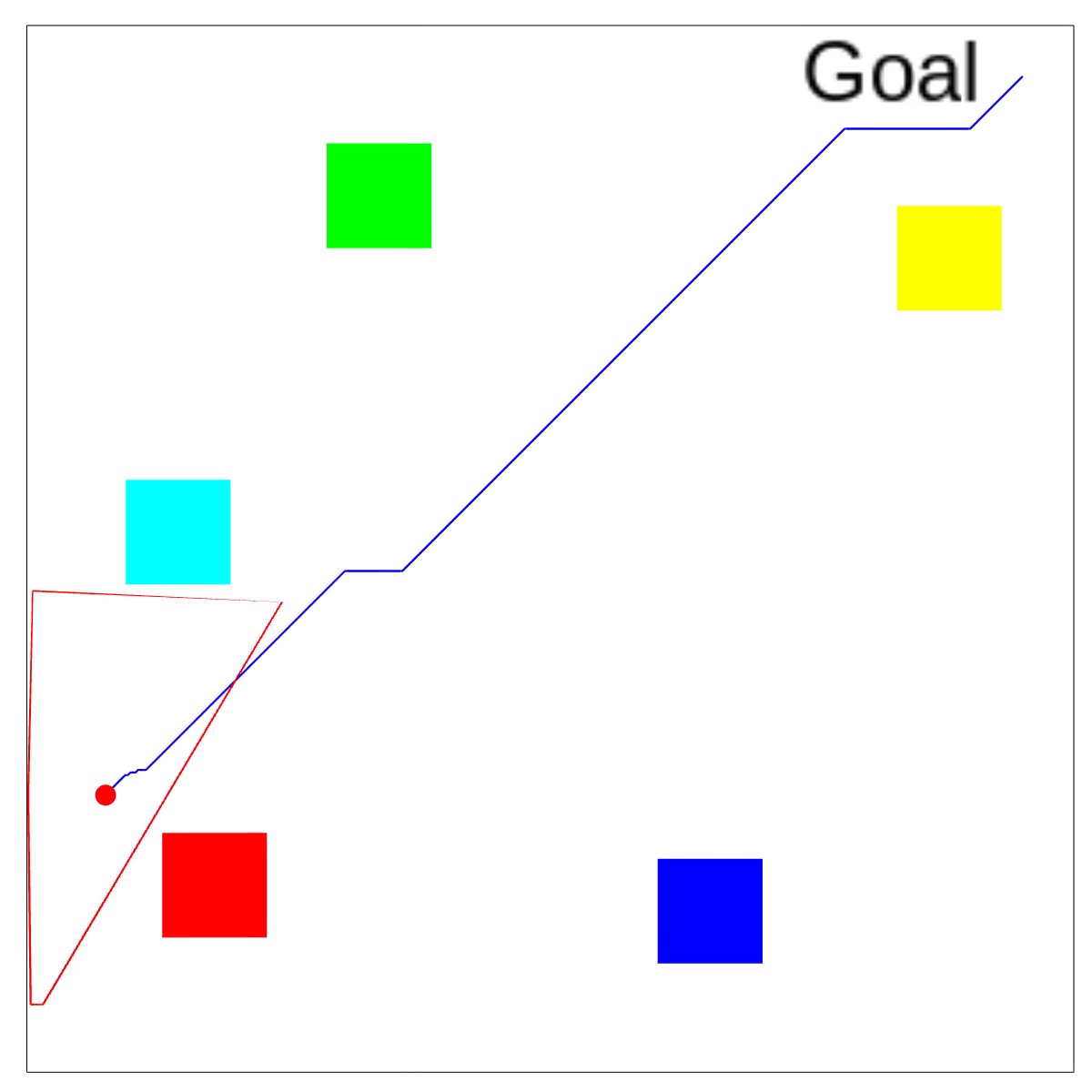}
\end{subfigure}
\begin{subfigure}{.160\textwidth}
  \centering
  \includegraphics[width=\linewidth]{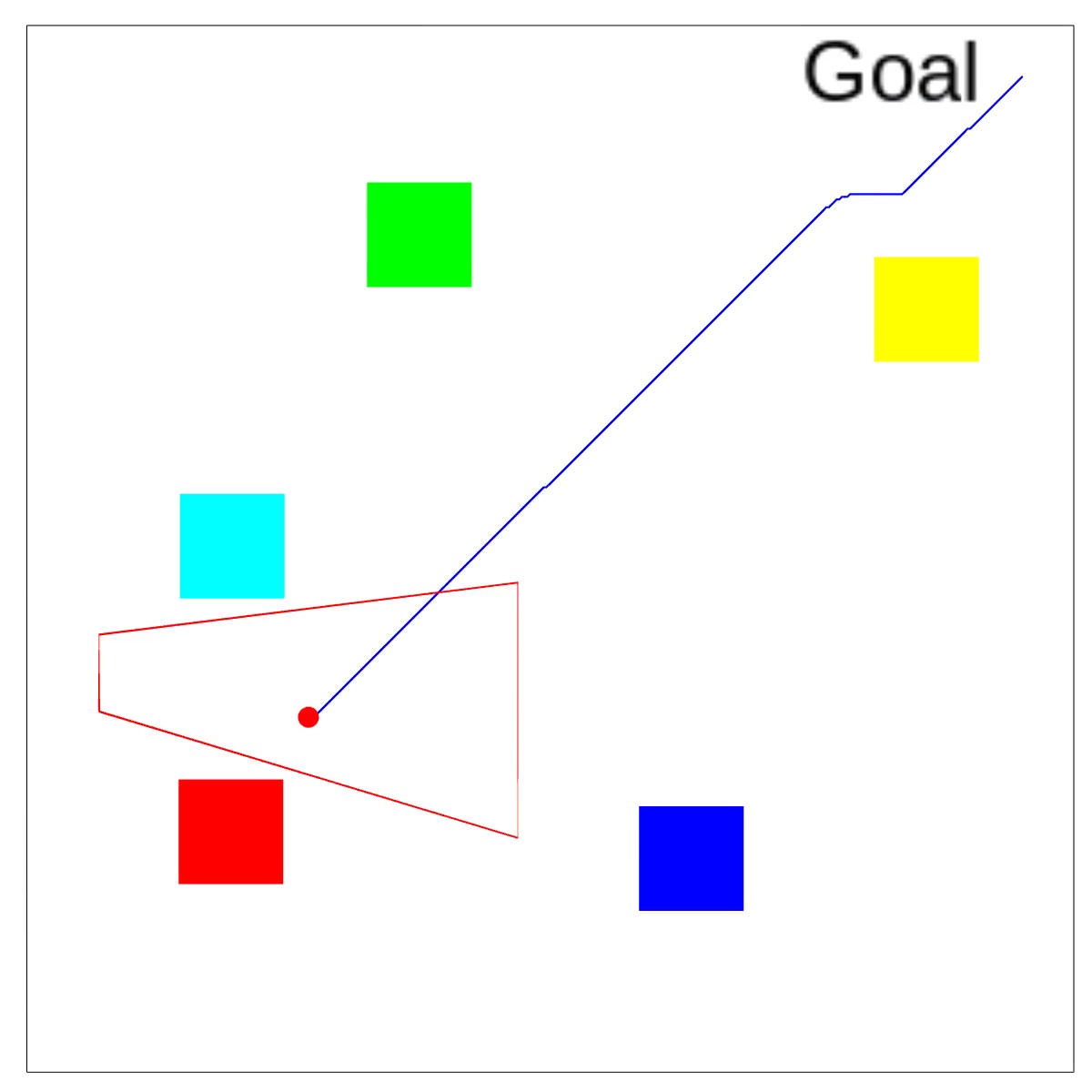}
\end{subfigure}
\begin{subfigure}{.160\textwidth}
  \centering
  \includegraphics[width=\linewidth]{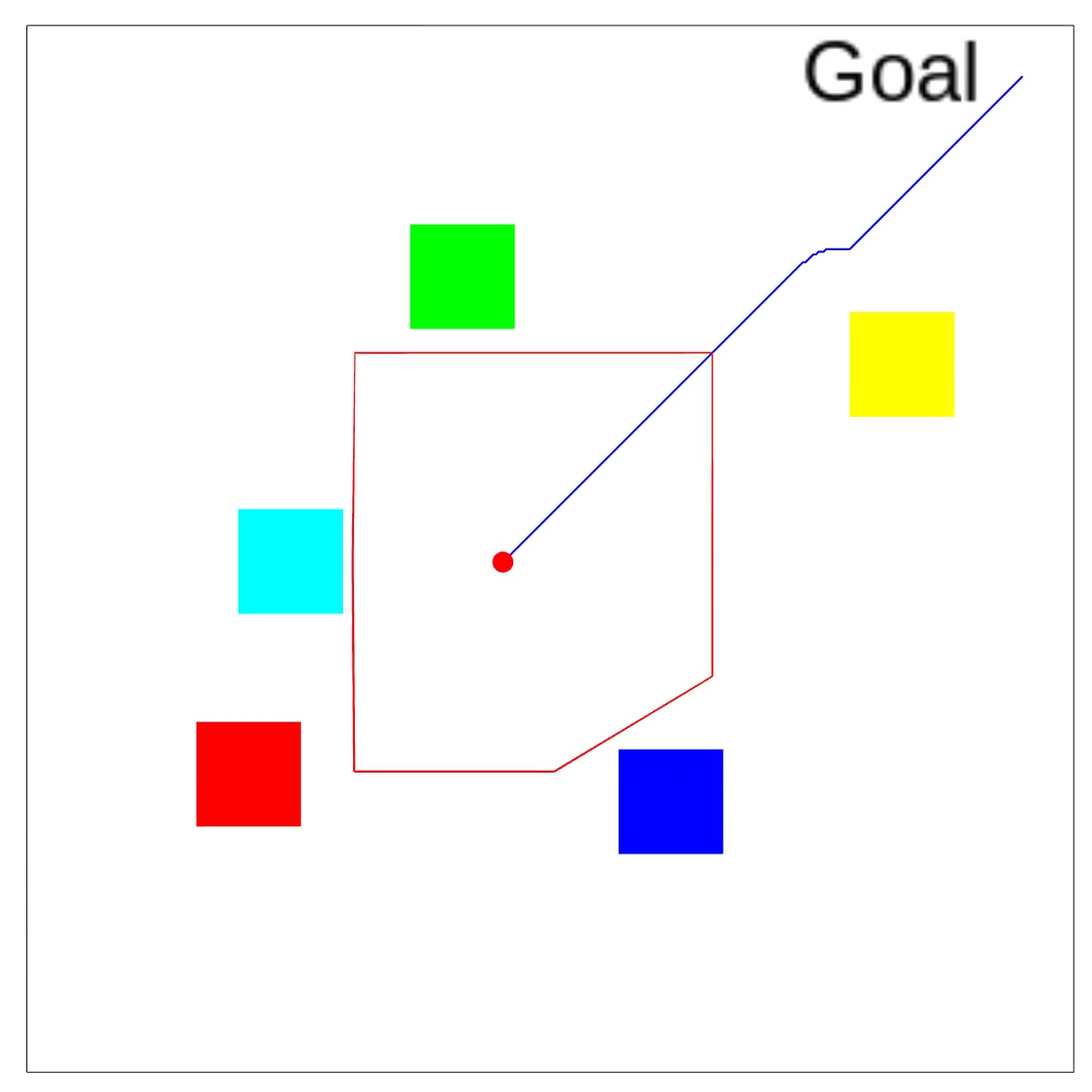}
\end{subfigure}
\begin{subfigure}{.160\textwidth}
  \centering
  \includegraphics[width=\linewidth]{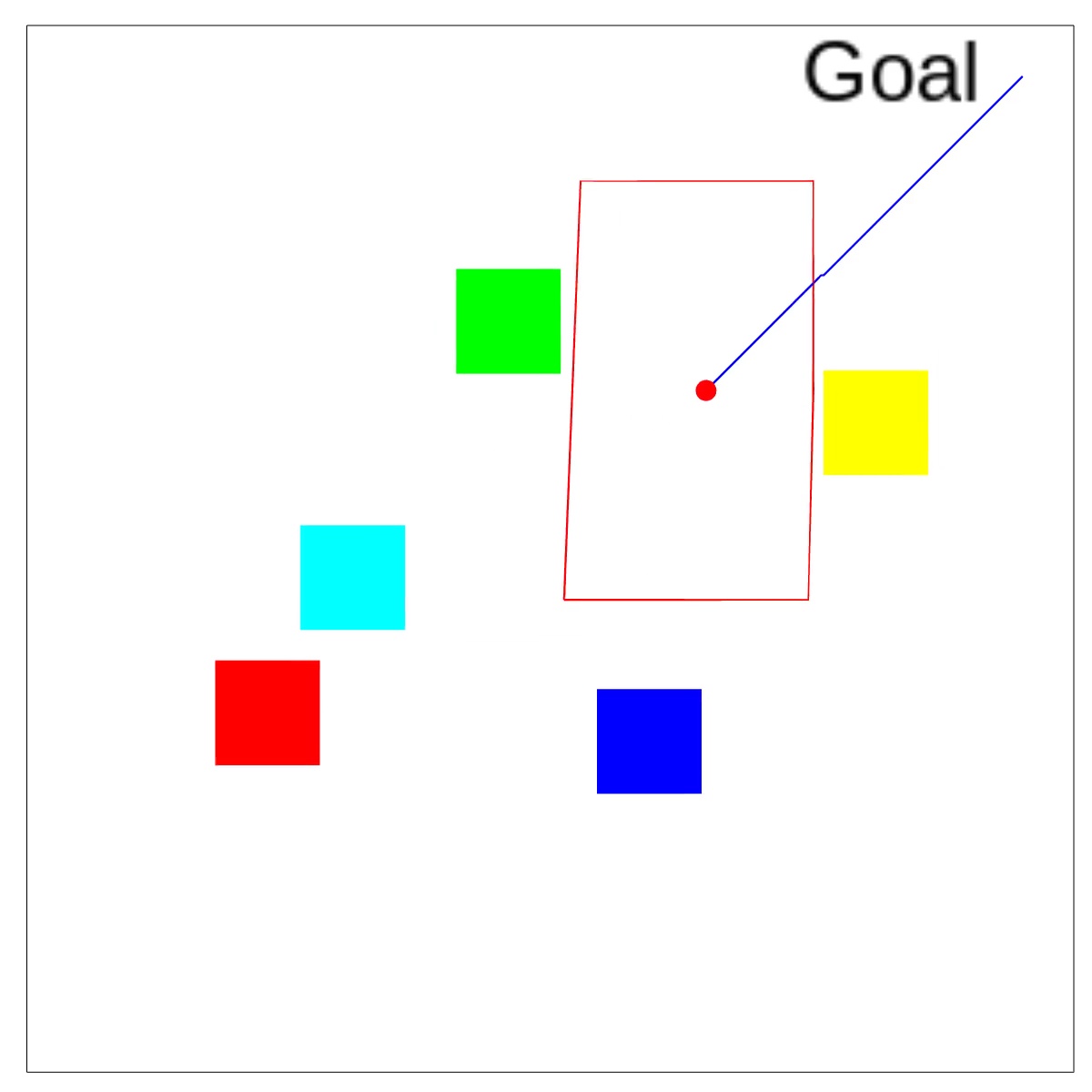}
\end{subfigure}
\begin{subfigure}{.160\textwidth}
  \centering
  \includegraphics[width=\linewidth]{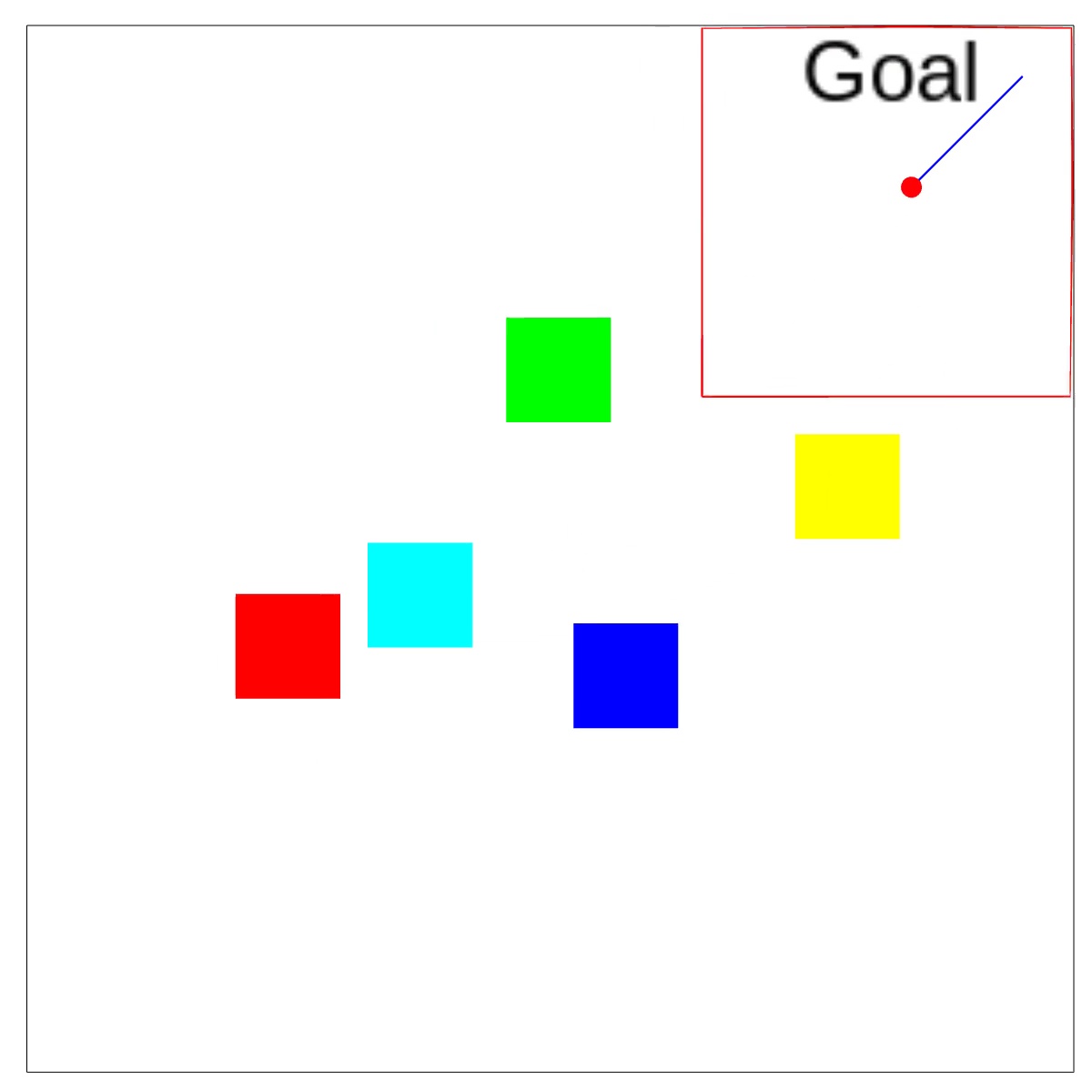}
\end{subfigure}
    \caption{Snapshots of the desired path (blue) and SCP (red polygons) at one-second intervals for avoiding 5 convex dynamic obstacles (colorful squares) with a controlled robot (small red circle), with horizon $N=30$. The robot safely reaches the goal location.
    } 
\label{fig:convex obstacles}
\end{figure*}

The 2-D map covers a square area from $[-50, -50]^T$ to $[50, 50]^T$, divided into 1200 by 1200 grid cells. We generated 5 dynamic square obstacles, each with a side length of 10 and a constant speed randomly chosen between $[6, 8]$. The robot, modeled as a circular unicycle with a radius of 1, avoids obstacles by inflating the obstacle boundaries by the robot's radius. Figure \ref{fig:convex obstacles} shows simulation snapshots where the robot successfully finds and follows a safe path. To evaluate computation speed and success rate, we varied the number and shapes of obstacles and the planning horizon, running 100 trials with random initial and target locations. Results in Tab. \ref{tab:compuation-time1} show that computation time increases with the number of obstacles and horizon length, while maintaining relatively fast processing (one-step computation time less than 55 milliseconds). The obstacle avoidance success rate exceeded $87\%$ for various obstacle configurations.

\begin{table*}[]
\centering
\resizebox{0.95\textwidth}{!}{
\begin{tabular}{|c|c|c|cccccc|}
\hline
\multicolumn{3}{|c|} {Convex obstacle shapes} & $n = 30$ & $n = 40$ & $n = 50$ &  $n = 60$ & $n = 70$  &  $n = 80$ \\ \hline
 \multirow{6}{*}{Square} & \multirow{2}{*}{N=10}  &  mean / std (ms)& $17.170\pm2.184$ & $20.260\pm2.382$ & $22.719\pm2.026$ & $26.304\pm2.238$ & $28.800\pm1.863$ & $32.146\pm2.211$ \\
& & feasibility rate & $97\%$ & $94\%$ & $90\%$ & $90\%$ & $89\%$ & $89\%$ \\ \cline{2-9} 
 & \multirow{2}{*}{N=20} & mean / std (ms) & $20.155\pm2.551$ & $22.546\pm2.461$ & $25.974\pm2.470$ & $29.073\pm2.044$ & $32.372\pm2.251$ & $34.993\pm2.188$ \\
 &  & feasibility rate &  $98\%$ & $93\%$ & $92\%$ & $93\%$ & $92\%$ & $89\%$ \\ \cline{2-9} 
 &\multirow{2}{*}{N=30} & mean / std (ms) & $22.365\pm2.109$ & $25.885\pm2.210$ & $28.960\pm2.352$ & $32.486\pm2.202$ & $36.438\pm2.281$ & $39.844\pm2.397$ \\
 & & feasibility rate &   $99\%$ & $96\%$ & $92\%$ & $92\%$ & $91\%$ & $90\%$ \\ \hline  
 
 \multirow{6}{*}{Triangle} & \multirow{2}{*}{N=10}  &  mean / std (ms)& $13.789\pm2.038$ & $15.713\pm2.210$ & $17.573\pm2.136$ & $19.073\pm2.322$ & $21.314\pm2.256$ & $22.699\pm1.768$ \\
& & feasibility rate &  $97\%$ & $96\%$ & $93\%$ & $95\%$ & $93\%$ & $93\%$ \\ \cline{2-9} 
 & \multirow{2}{*}{N=20} & mean / std (ms) & $16.640\pm2.251$ & $18.726\pm1.950$ & $21.692\pm2.313$ & $24.173\pm2.391$ & $26.449\pm2.271$ & $28.730\pm2.145$ \\
 &  & feasibility rate &   $98\%$ & $98\%$ & $97\%$ & $98\%$ & $95\%$ & $93\%$ \\ \cline{2-9} 
 &\multirow{2}{*}{N=30} & mean / std (ms) & $20.755\pm2.236$ & $23.641\pm2.611$ & $26.250\pm2.346$ & $29.405\pm2.463$ & $32.027\pm2.464$ & $34.893\pm2.224$ \\
 & & feasibility rate &   $98\%$ & $98\%$ & $95\%$ & $94\%$ & $95\%$ & $95\%$ \\ \hline  

 \multirow{6}{*}{Circle} & \multirow{2}{*}{N=10}  &  mean / std (ms)& $18.145\pm2.222$ & $19.175\pm1.996$ & $22.165\pm1.880$ & $25.100\pm2.104$ & $27.608\pm1.992$ & $30.577\pm1.811$ \\
& & feasibility rate & $97\%$ & $97\%$ & $94\%$ & $94\%$ & $93\%$ & $89\%$ \\  \cline{2-9} 
 & \multirow{2}{*}{N=20} & mean / std (ms) & $23.133\pm2.273$ & $26.949\pm2.232$ & $32.218\pm2.323$ & $35.631\pm2.728$ & $40.127\pm2.574$ & $42.838\pm2.745$ \\
 &  & feasibility rate &  $96\%$ & $97\%$ & $96\%$ & $95\%$ & $91\%$ & $90\%$ \\  \cline{2-9} 
 &\multirow{2}{*}{N=30} & mean / std (ms) & $27.579\pm2.496$ & $33.309\pm2.262$ & $40.488\pm2.414$ & $44.036\pm2.256$ & $49.365\pm2.634$ & $51.857\pm2.248$ \\
 & & feasibility rate & $99\%$ & $94\%$ & $93\%$ & $90\%$ & $90\%$ & $87\%$ \\ \hline
\end{tabular}
}
\caption{Statistical benchmarks for computation times and feasibility rates of our approach with randomized states. $N$ is the horizon length and $n$ is the number of obstacles. The one-step computation time is reported by mean/standard deviation over 100 trails, and the feasibility rate is the number of times the robot successfully reaches the goal location. More information can be found at \url{https://youtu.be/iiihs_KPEZQ}.
\label{tab:compuation-time1}}
\end{table*}

\begin{table*}[]
\centering
\resizebox{0.95\textwidth}{!}{
\begin{tabular}{|c|c|c|cccccc|}
\hline
\multicolumn{3}{|c|} {Nonconvex obstacle shapes} & $n = 4$ & $n = 5$ & $n = 6$ &  $n = 7$ & $n = 8$  &  $n = 9$ \\ \hline
 \multirow{6}{*}{3 blades} & \multirow{2}{*}{N=10}  &  mean / std (ms)& $12.078\pm2.095$ & $13.336\pm2.174$ & $14.325\pm2.074$ & $15.327\pm2.201$ & $16.343\pm2.091$ & $17.549\pm2.079$ \\
& & feasibility rate & $99\%$ & $97\%$ & $95\%$ & $95\%$ & $88\%$ & $90\%$ \\ \cline{2-9} 
 & \multirow{2}{*}{N=20} & mean / std (ms) & $14.479\pm2.102$ & $15.800\pm2.139$ & $17.423\pm2.238$ & $18.531\pm2.121$ & $20.326\pm2.188$ & $21.373\pm2.07$ \\
 &  & feasibility rate &  $97\%$ & $90\%$ & $93\%$ & $93\%$ & $93\%$ & $89\%$ \\ \cline{2-9} 
 &\multirow{2}{*}{N=30} & mean / std (ms) & $17.728\pm2.137$ & $19.210\pm2.285$ & $20.613\pm2.333$ & $22.613\pm2.152$ & $23.990\pm2.631$ & $26.191\pm2.465$ \\
 & & feasibility rate &   $97\%$ & $98\%$ & $96\%$ & $93\%$ & $93\%$ & $89\%$ \\ \hline  
 
 \multirow{6}{*}{4 blades} & \multirow{2}{*}{N=10}  &  mean / std (ms)&  $15.062\pm2.135$ & $16.761\pm2.221$ & $18.848\pm2.212$ & $20.315\pm2.346$ & $22.360\pm2.076$ & $23.736\pm2.210$ \\
& & feasibility rate &  $97\%$ & $97\%$ & $98\%$ & $94\%$ & $92\%$ & $87\%$ \\ \cline{2-9} 
 & \multirow{2}{*}{N=20} & mean / std (ms) & $17.500\pm2.203$ & $19.731\pm2.099$ & $21.897\pm2.166$ & $23.903\pm2.149$ & $26.240\pm2.290$ & $28.187\pm2.360$ \\
 &  & feasibility rate & $97\%$ & $97\%$ & $97\%$ & $94\%$ & $95\%$ & $91\%$ \\  \cline{2-9} 
 &\multirow{2}{*}{N=30} & mean / std (ms) & $21.097\pm2.290$ & $23.249\pm2.412$ & $25.613\pm2.364$ & $28.717\pm2.404$ & $31.339\pm2.478$ & $33.420\pm2.490$ \\
 & & feasibility rate &  $97\%$ & $97\%$ & $97\%$ & $95\%$ & $97\%$ & $97\%$ \\  \hline  

 \multirow{6}{*}{5 blades} & \multirow{2}{*}{N=10}  &  mean / std (ms)&  $18.152\pm2.070$ & $21.379\pm2.148$ & $23.715\pm2.148$ & $26.664\pm2.290$ & $29.130\pm2.271$ & $31.659\pm2.171$ \\
& & feasibility rate & $98\%$ & $96\%$ & $96\%$ & $94\%$ & $94\%$ & $95\%$ \\  \cline{2-9} 
 & \multirow{2}{*}{N=20} & mean / std (ms) &  $21.830\pm2.481$ & $24.514\pm2.139$ & $27.634\pm2.600$ & $30.272\pm2.437$ & $33.622\pm2.504$ & $36.545\pm2.433$ \\
 &  & feasibility rate &  $96\%$ & $94\%$ & $95\%$ & $95\%$ & $95\%$ & $94\%$ \\  \cline{2-9} 
 &\multirow{2}{*}{N=30} & mean / std (ms) & $24.773\pm2.378$ & $27.947\pm2.358$ & $32.266\pm2.735$ & $35.513\pm2.727$ & $39.214\pm2.798$ & $42.290\pm2.898$ \\
 & & feasibility rate &  $98\%$ & $99\%$ & $98\%$ & $98\%$ & $95\%$ & $94\%$ \\ \hline
\end{tabular}
}
\caption{Computation time and feasibility rates of our approach for nonconvex obstacles (fan-shaped obstacles with 3, 4, and 5 blades). More information can be found at \url{https://youtu.be/iiihs_KPEZQ}.
\label{tab:compuation-time2}}
\end{table*}

\begin{table}[]
\vspace{3mm}
    \centering
    \resizebox{0.49\textwidth}{!}{
    \begin{tabular}{|c|cccc|}
    \hline
        Approaches & Traj. length (m) & Comp. time (s)& Step count & Feas. rate\\
    \hline
        CBF-RRT & $7.324\pm0.601$ & $12.244\pm5.760$ & $36.720\pm3.004$ & $94\%$\\
    \hline
        CBF-CLF-1 &  $9.410\pm8.202$ & $0.070\pm0.061$  & $51.638\pm45.530$& $94\%$ \\
    \hline
        CBF-CLF-2 &  $7.958\pm6.272$ & $0.060\pm0.050$  & $43.622\pm34.912$& $74\%$ \\
    \hline
        iMPC-CBF-JPS & $6.218\pm0.742$ & $0.285\pm0.094$ & $31.430\pm4.732$  & $100\%$  \\ 
    \hline
    \end{tabular}}
    \caption{Comparison of overall trajectory length, total computation time, total time steps, and feasibility rate for the three methods. All data are based on the mean and standard deviation over 50 trials. The horizon length for iMPC-(DHO)CBF-JPS is 10.}
    \label{tab:comparison3}
\end{table}

\subsubsection{Avoiding Nonconvex Dynamic Obstacles}
\begin{figure*}[t]
\centering
\begin{subfigure}{.160\textwidth}
  \centering
  \includegraphics[width=\linewidth]{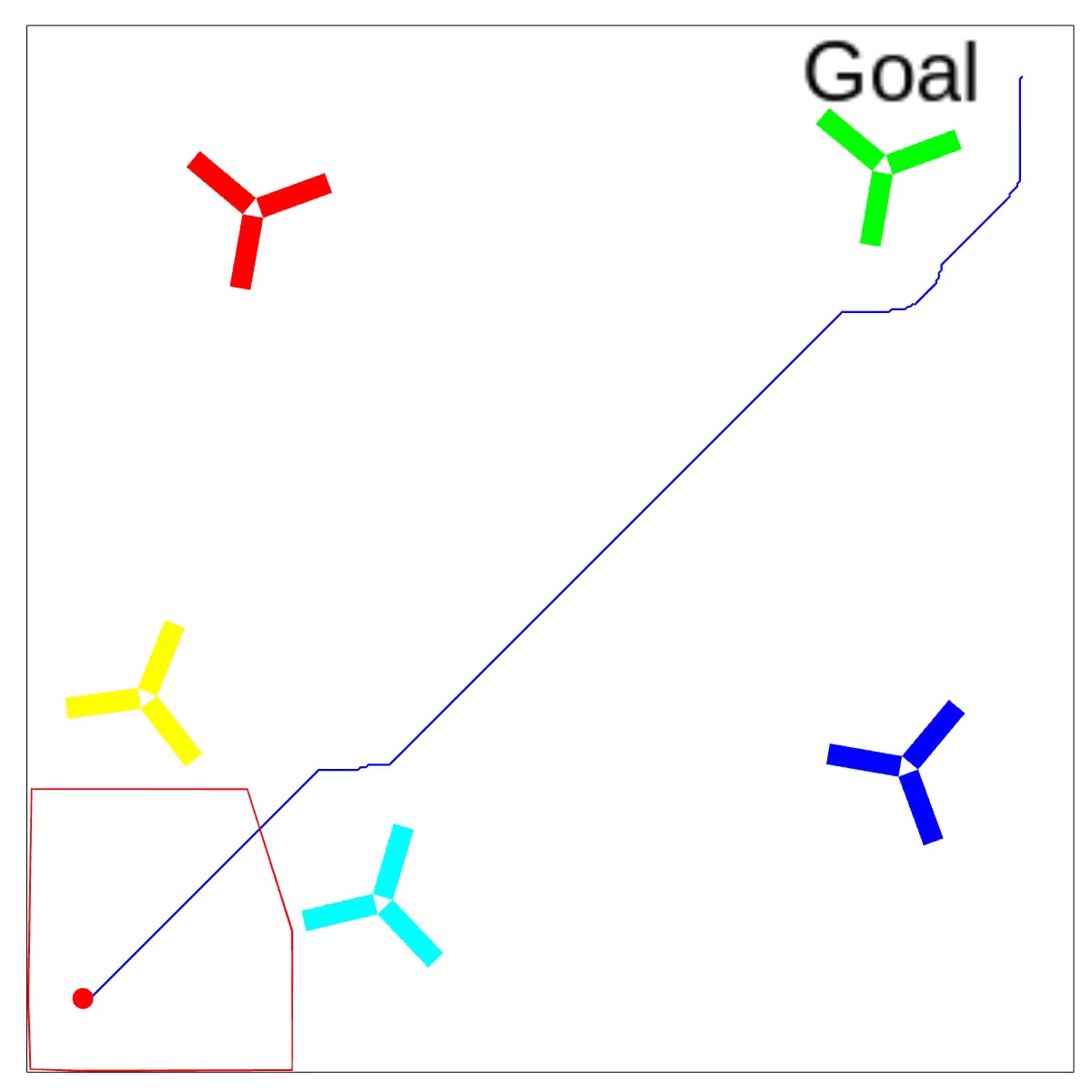}
\end{subfigure}
\begin{subfigure}{.160\textwidth}
  \centering
  \includegraphics[width=\linewidth]{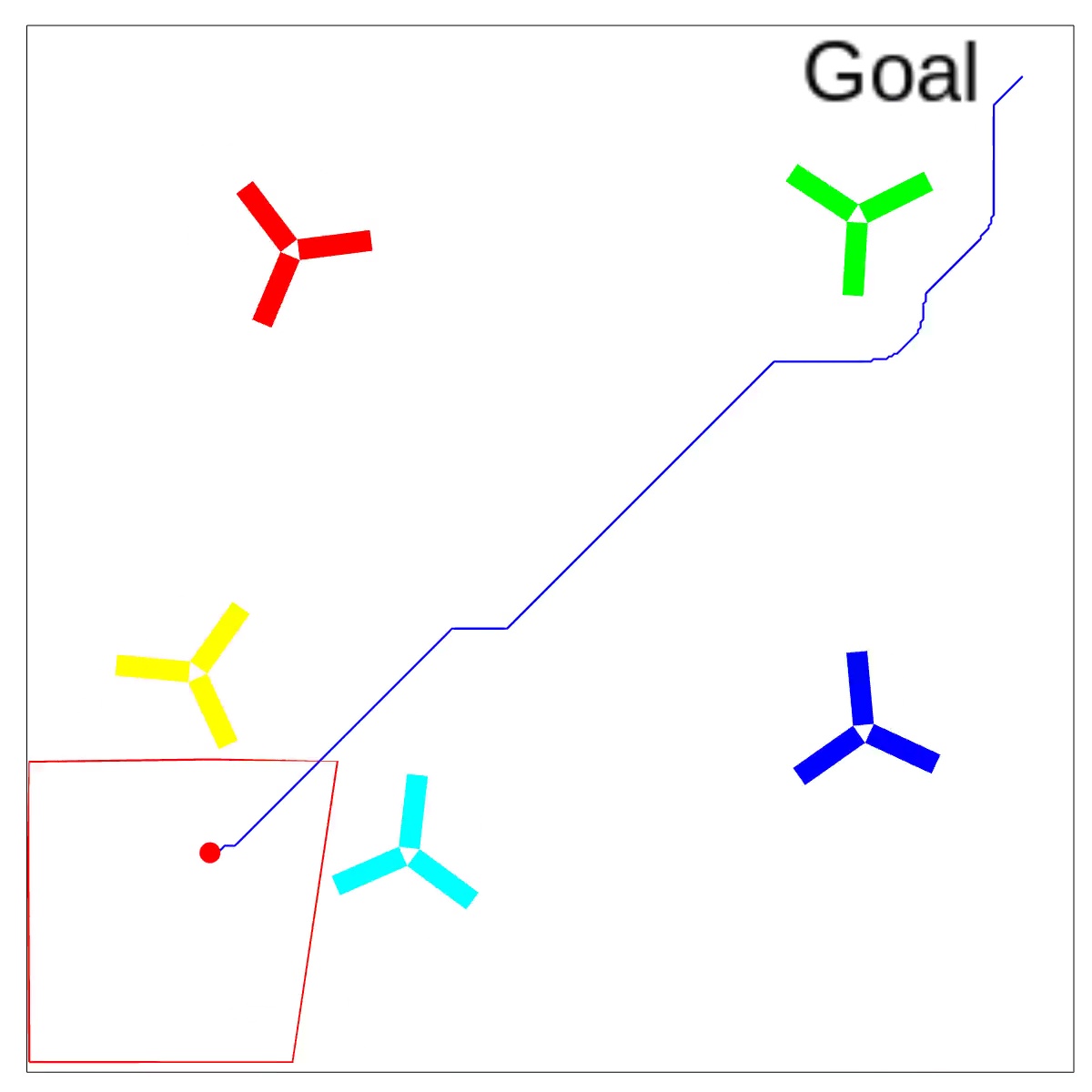}
\end{subfigure}
\begin{subfigure}{.160\textwidth}
  \centering
  \includegraphics[width=\linewidth]{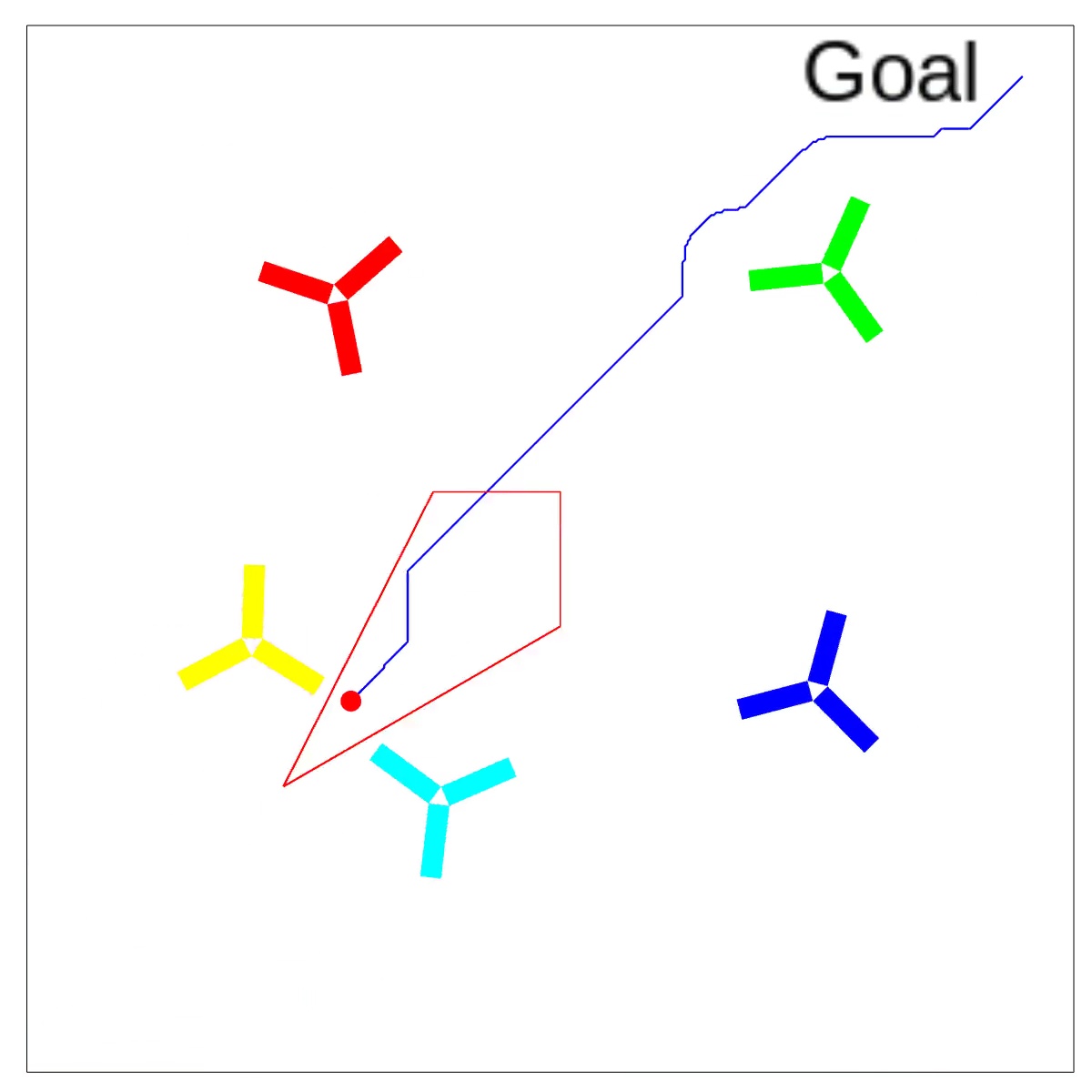}
\end{subfigure}
\begin{subfigure}{.160\textwidth}
  \centering
  \includegraphics[width=\linewidth]{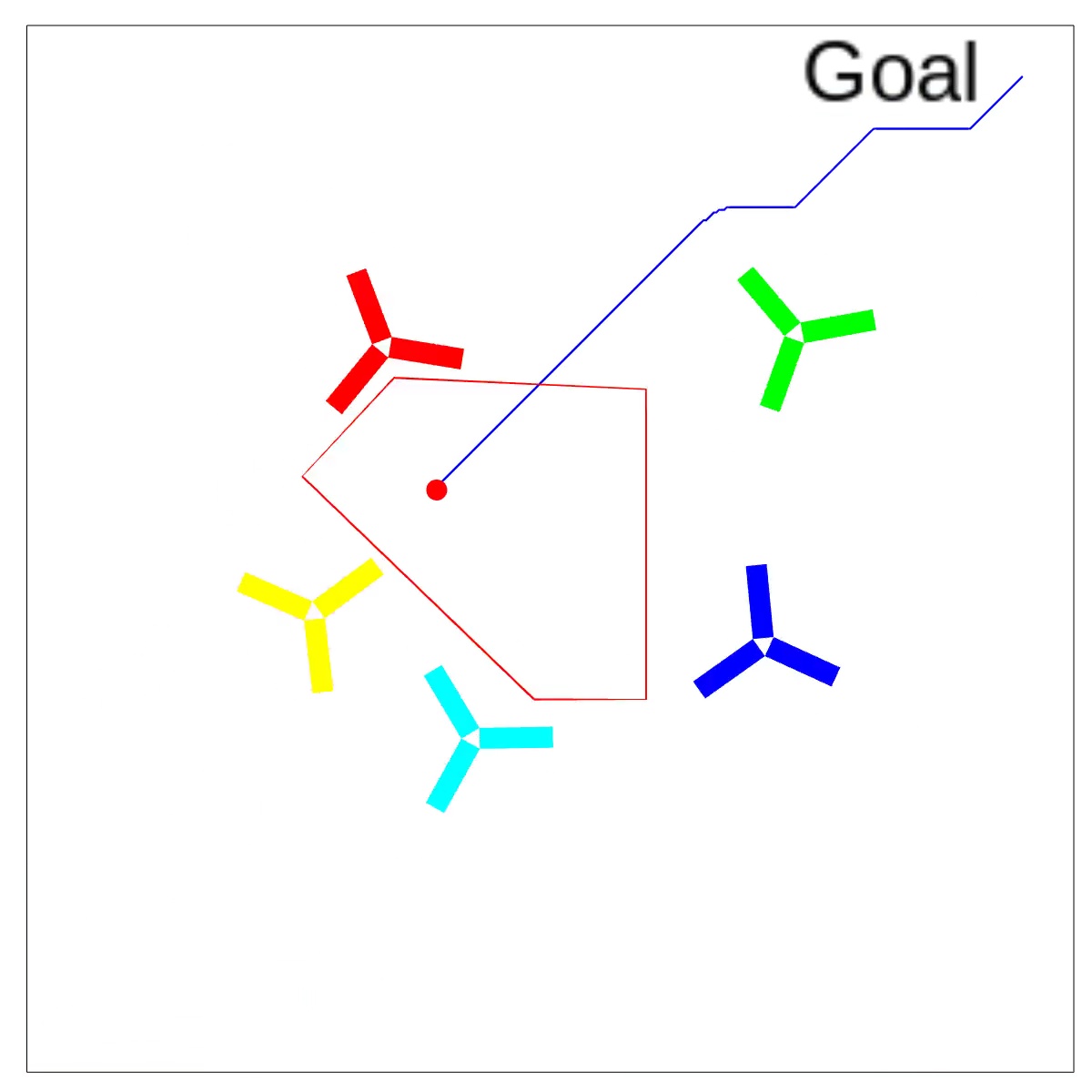}
\end{subfigure}
\begin{subfigure}{.160\textwidth}
  \centering
  \includegraphics[width=\linewidth]{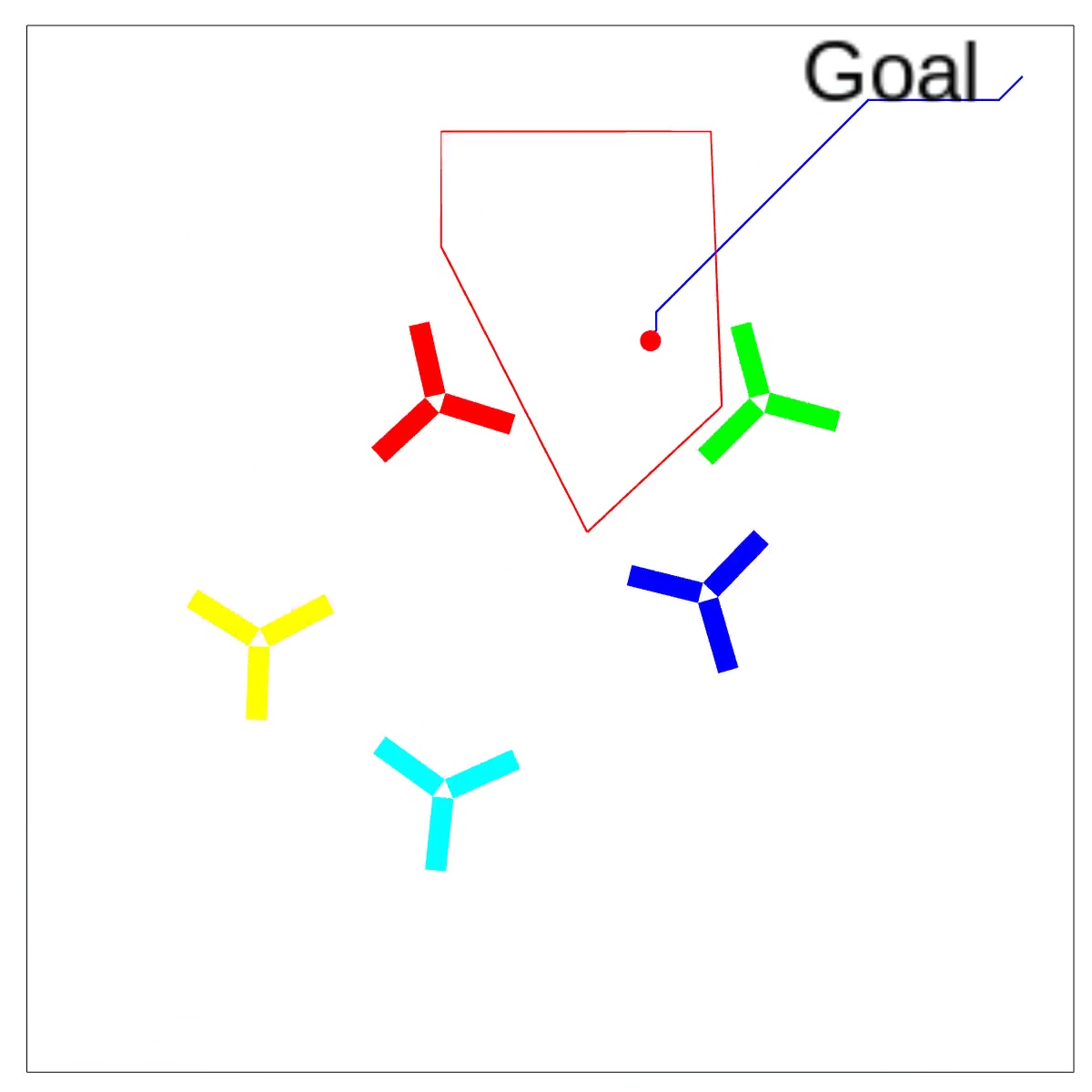}
\end{subfigure}
\begin{subfigure}{.160\textwidth}
  \centering
  \includegraphics[width=\linewidth]{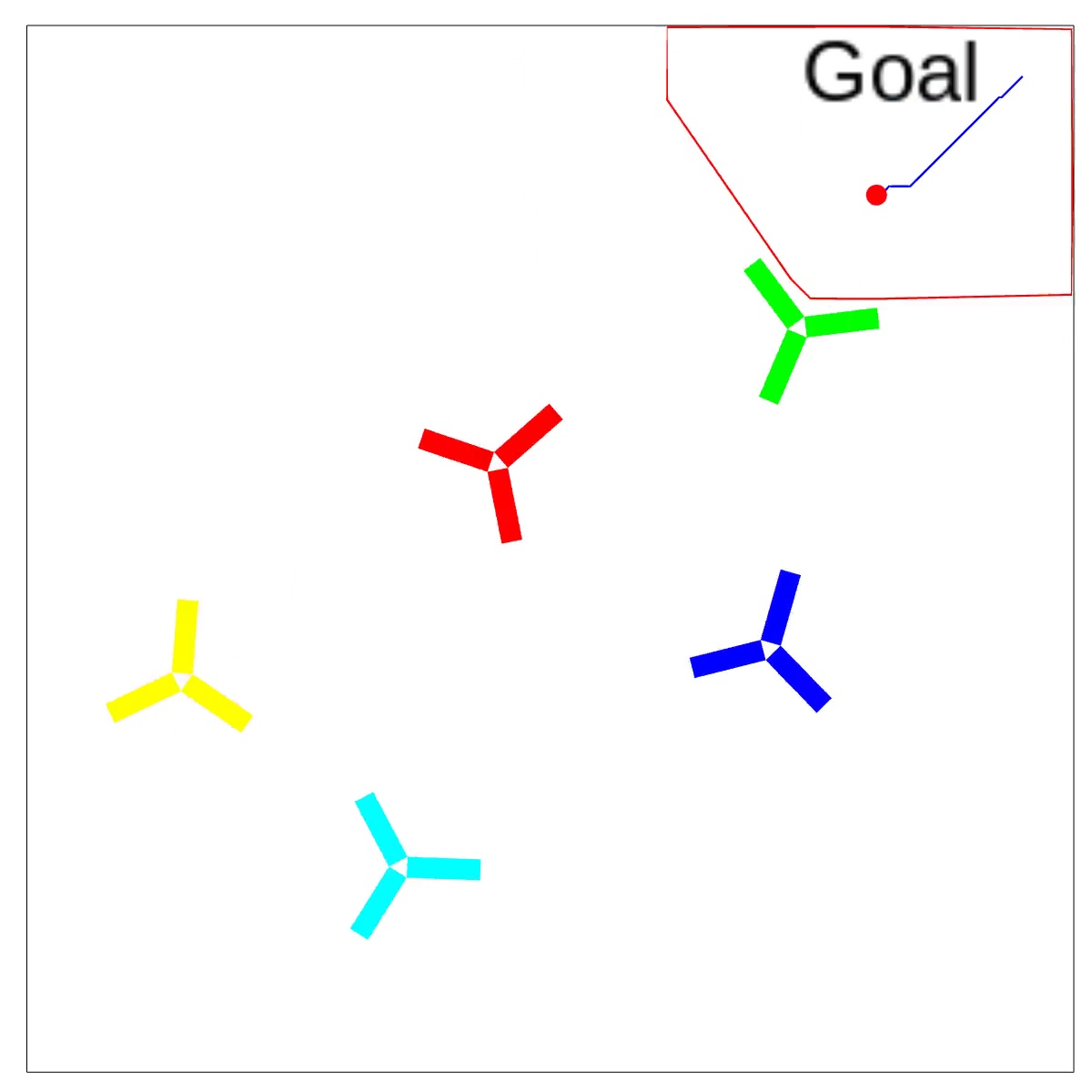}
\end{subfigure}
    \caption{Snapshots of desired path (blue) and SCP (red polygons) at one-second intervals for avoiding 5 nonconvex-shaped dynamic obstacles (colorful fans) with a controlled robot (small red circle), horizon $N=30$. The robot can safely reach the goal location.
    } 
\label{fig:nonconvex obstacles}
\end{figure*}

To demonstrate our algorithm's effectiveness in a more complex map, we generated 5 rotating fans, each with 3 equidistant blades of length 7 and width 2. The blades' geometric centers have a constant translational speed randomly chosen within $[6, 8]$, and the rotational speed is randomly set within $[2, 8]$. Fig. \ref{fig:nonconvex obstacles} shows that the robot can still find and follow a safe path during the simulation. As in the previous experiment, we increased the number of obstacles and blades and varied the horizon length. The results in Tab. \ref{tab:compuation-time2} indicate that the computation time grows with the number of obstacles and the length of the horizon, but remains relatively fast (one-step computation time less than 50 milliseconds), with an obstacle avoidance success rate over $87\%$ across different configurations.

\subsubsection{Comparison with Benchmarks}
\begin{figure}[ht]
    \centering
    \includegraphics[scale=0.46]{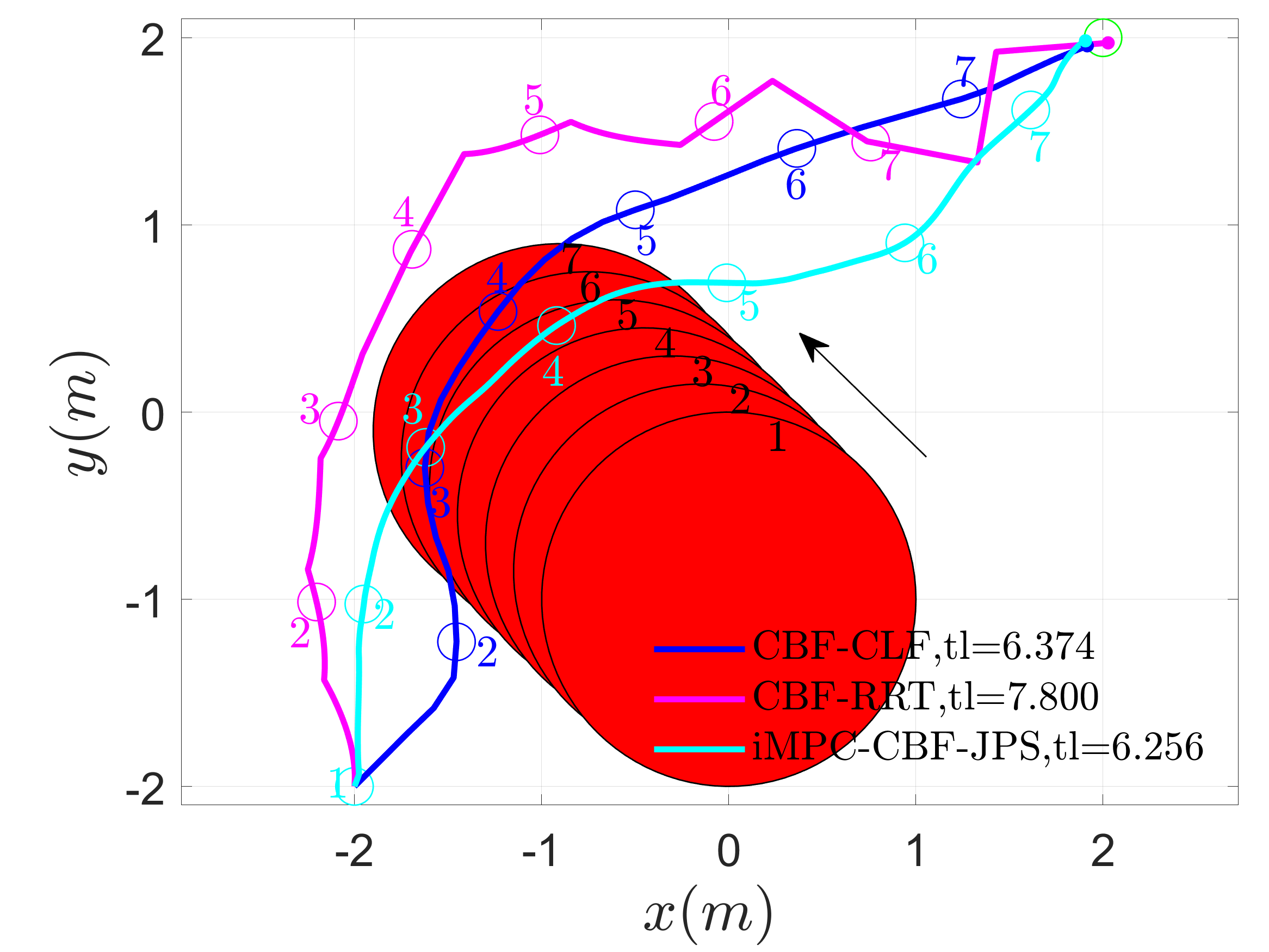}
    \caption{Trajectory comparison for the three approaches for avoiding a circular dynamic obstacle: tl is the overall trajectory length, and the goal is the green circle. Seven robot locations (marked by circles) and obstacle locations (arrow shows its moving direction) are shown at half-second intervals. All three approaches ensure safe navigation, with proposed method achieving the shortest trajectory.}
    \label{fig:comparison3}
\end{figure}

We selected CBF-CLF from \cite{huang2023obstacle} and CBF-RRT from \cite{yang2019sampling} as benchmarks for the simplified unicycle model \eqref{eq:simplified-unicycle-model} to avoid a circular dynamic obstacle. The radius of the robot is 0.1. For fair comparison, all methods start from $[-2,-2]$ with a heading angle of $\frac{\pi}{4}$, targeting $[2,2]$, with a fixed linear speed of 2, $\Delta t = 0.1$, and $T = 300$. For the benchmarks, the CBF candidates are defined by the Euclidean distance between the center of the robot and the obstacle (the authors of \cite{huang2023obstacle} used the rear axle axis to represent the robot's center coordinates, while the authors of \cite{yang2019sampling} considered only the center position of the robot. Consequently, the relative degree of the CBF candidate is 1 in \cite{huang2023obstacle} but 2 in \cite{yang2019sampling}). 
 The values for the hyperparameters are takes from the respective papers, except for $\alpha(h(\mathbf{x})) = 8h(\mathbf{x})$ for CBF-CLF. The obstacle, with a radius of 1, moves from $[0, -1]$ at a speed of $[-0.3, -0.3]$, where the first component is the x-axis speed and the second is the y-axis speed. As shown in Fig. \ref{fig:comparison3}, our iMPC-CBF-JPS provides a safer control strategy than CBF-CLF and a shorter trajectory than both CBF-CLF and CBF-RRT. We ran 50 experiments with varying obstacle locations and speeds, as summarized in Tab. \ref{tab:comparison3}. For CBF-CLF-1, $\alpha(h(\mathbf{x})) = 8h(\mathbf{x})$ and for CBF-CLF-2, $\alpha(h(\mathbf{x})) = 10h(\mathbf{x})$. The results indicate that increasing the coefficient for CBF-CLF reduces trajectory length but lowers success rates. CBF-RRT shows slower computation with success rates similar to CBF-CLF. Our method achieves the shortest trajectory and the highest success rate, with computation times slightly longer than CBF-CLF, while still maintaining fast per-step computation speeds.

\section{Conclusion and Future Work}
\label{sec:Conclusion and Future Work}
We proposed an iterative convex optimization procedure using MPC for dynamic obstacle avoidance in grid maps. In this framework, we first obtain discrete-time high-order CBFs from convex polytopes to ensure safety requirements, without needing to know the boundary equations of the obstacles.  We validated our approach on navigation problems with obstacles of varying numbers, speeds, and shapes, and showed that our method outperforms the state of the art. We are currently working on predicting future information about dynamic obstacles in our framework. 

\bibliographystyle{IEEEtran}
\balance
\bibliography{references.bib}

\end{document}